\documentclass{article}

\usepackage[a4paper, total={6in, 8in}]{geometry}

\usepackage{natbib}
 \bibpunct[, ]{(}{)}{,}{a}{}{,}%

\usepackage{algorithm}
\usepackage{algpseudocode}
\usepackage{tikz}
\usetikzlibrary{positioning,fit,backgrounds, calc,arrows.meta}
\usepackage{graphicx}
\usepackage{caption}
\usepackage{subcaption}
\usepackage{url}
\usepackage{amssymb,amsmath,amsthm}
\usetikzlibrary{shapes.geometric}
\usepackage{xparse}
\usepackage{pgfplots}
\usepackage{pgfplotstable}
\pgfplotsset{compat=1.18}

\def\argmax{\mathop{\rm arg\,max}}%
\def\argmin{\mathop{\rm arg\,min}}%

\pgfplotsset{myCurveStyle/.style={thick, mark=none}}

\definecolor{viridis1}{RGB}{68,1,84}
\definecolor{viridis2}{RGB}{58,82,139}
\definecolor{viridis3}{RGB}{32,144,140}
\definecolor{viridis4}{RGB}{94,201,97}
\definecolor{viridis5}{RGB}{253,231,36}

\pgfplotsset{
    mySimAxisStyle/.style={
        xlabel={Training Data Size ($250\times 2^x$)},
        ylabel={Improvement (\%)},
        xmin=-0.25, xmax=12.25,
        grid=major,
        scale only axis=true,
        scale=0.82,
        legend pos=south east,
        legend style={anchor=south, font=\footnotesize},
        cycle list={{blue},{red},{green!50!black},{orange}},
    },
    SimCVAxisStyle/.style={
        mySimAxisStyle,
        ylabel={Improvement (\%)},
        legend pos=north east,
        legend style={anchor=north, font=\footnotesize},
        cycle list={{viridis1}, {viridis2}, {viridis3}, {viridis4}, {viridis5}},
    },
}

\pgfplotsset{
    myCriteoAxisStyle/.style={
        xlabel={Training Data Size (in 10,000s)},
        ylabel={Improvement (\%)},
        xtick={1, 4, ..., 25},
        xmin=0.5, xmax=25.5,
        grid=major,
        scale only axis=true,
        scale=0.82,
        legend pos=south east,
        legend style={anchor=south, font=\footnotesize},
        cycle list={{blue},{red},{green!50!black},{orange}},
    },
}

\pgfplotsset{
  colormap={MyAccent}{
    rgb255=(127, 201, 127)
    rgb255=(190, 174, 212)
    rgb255=(191, 91, 22)
    rgb255=(255, 255, 153)
    rgb255=(56, 108, 176)
    rgb255=(102, 102, 102) 
  }
}

\pgfplotsset{
    myHeatmapStyle/.style={
        axis on top, 
        ymin=-0.5, ymax=4.5,
        ytick={0,1,...,4},
        yticklabels={0.1, 0.3, 0.5, 0.7, 0.9},
        ylabel={Alignment ($\rho$)},
        major tick length=0pt,
        enlarge x limits=false, 
        enlarge y limits=false,
        colormap name=MyAccent,
        colorbar style={hidden},
        point meta min=0,
        point meta max=5,
        scale only axis=true,
        scale=0.88
    },
    HeatmapNoiseFixed/.style={
        myHeatmapStyle, 
        xmin=-0.5, xmax=8.5,
        xtick={0,1,...,8},
        xticklabels={0,1,2,3,4,5,6,7,8},
        xlabel={Training Data Size ($250\times 2^x$)},
    },
    HeatmapDataSizeFixed/.style={
        myHeatmapStyle,
        xmin=-0.5, xmax=6.5,
        xtick={0,1,...,6},
        xticklabels={6.4,3.2,1.6,0.8,0.4,0.2,0.1},
        xlabel={Variance of Noise ($\sigma_{\epsilon}^2$)},
    }
}

\ExplSyntaxOn
\NewExpandableDocumentCommand{\getLabel}{m}
{
 \str_case:nnF { #1 }
  {
   {BS}   {BS}
   {MC}    {P-CAL}
   {CF}    {CM}
   {CF-BS}  {P-COV}
   {CRF}   {P-RES}
  }
  { #1 } 
}
\ExplSyntaxOff

\begin{document}




\title{Causal Post-Processing of Predictive Models}

\author{
{Carlos Fern\'andez-Lor\'ia}\\
Hong Kong University of Science and Technology\\
\url{imcarlos@ust.hk}\\ \\
{Yanfang Hou}\\ 
Hong Kong University of Science and Technology\\
\url{yanfang.hou@connect.ust.hk}\\ \\
{Foster Provost}\\
New York University\\
\url{fprovost@stern.nyu.edu}\\ \\
{Jennifer Hill}\\
New York University\\
\url{jennifer.hill@nyu.edu}}

\date{}
\maketitle

\begin{abstract}
Organizations increasingly rely on predictive models to decide who should be targeted for interventions—such as marketing campaigns, customer retention offers, or medical treatments. Yet these models are usually built to predict outcomes (e.g., likelihood of purchase or churn), not the actual impact of an intervention. As a result, the scores they produce---their predicted values---are often imperfect guides for allocating resources. Causal effects can be estimated with randomized experiments, but experiments are costly, limited in scale, and tied to specific actions. We propose \emph{causal post-processing} (CPP), a family of techniques that uses limited experimental data to refine the outputs of predictive models, so they better align with causal decision making. The CPP family spans approaches that trade off flexibility against data efficiency, unifying existing methods and motivating new ones. Through simulations and an empirical study in digital advertising, we show that CPP can improve intervention decisions, particularly when predictive models capture a useful but imperfect causal signal. Our results show how organizations can combine predictive modeling with experimental evidence to make more effective and scalable intervention decisions.
\end{abstract}


\maketitle

%


\section{Introduction}

Organizations routinely use predictive models to decide \textit{who to intervene on}, with the goal of influencing outcomes. What is striking is that these models are not designed to estimate causal effects---yet the scores they produce are used as proxies for causal impact and can be surprisingly effective in targeting interventions.

This practice is common. Marketing models used for targeting often predict the probability a consumer will purchase, rather than the causal effect on purchase~\citep{stitelman2011estimating}.  Moreover, due to the dearth of data on purchases, proxies often are used.  For example, models predict the probability that a consumer will visit a brand's website, and that probability is treated as a signal for how much an ad might sway purchasing~\citep{dalessandro2015evaluating}.  Similarly, retention models for targeting incentive offers normally estimate churn likelihood, not the causal effect on retention~\citep{ascarza2018pursuit}. Recommender systems are often based on predictions of what users would select on their own, yet they are deployed with the goal of \textit{influencing} those choices to increase engagement or purchases~\citep{li2022recommender, lee2020different}.

Two reasons combine to justify employing these imperfect proxies.   First, building causal models to estimate heterogeneous treatment effects can be very expensive.  We will return to that shortly.  Second, these proxy models can work, because their estimands often correlate well with the causal effect of interest.  In fact, effective decisions often do not require causal estimates, even when the intent is to causally influence outcomes~\citep{fernandez2022causal}. In many cases, the objective is to prioritize---ranking individuals so that scarce incentive resources go where they are most effective. In other cases, the goal is ``causal classification'': selecting a subset (e.g., the top 10\% of customers) for treatment, without requiring perfect ordering within that group. For such ranking and classification tasks, proxy scores are inexpensive to obtain, widely available, and often surprisingly effective. 

The fundamental challenge, of course, is that proxy scores are not causal effect estimates. A churn model, for example, predicts risk---not how a retention offer would reduce the risk. Targeting strategies based on such non-causal scores may still work well if the proxy is well correlated with causal impact, but they risk being misaligned when it is not.

The obvious alternative is to estimate heterogeneous causal effects directly. Randomized experiments can provide data to make this possible, revealing how individuals behave with and without treatment.  If there is enough training data, we can build models to estimate individual-level effects. 

The drawback is that experimentation is costly. It requires random targeting, which means withholding interventions from some who might benefit and giving them to others who might not. When proxy scores already provide a strong basis for decisions, the opportunity cost of randomization can be substantial. In some settings, experimentation platforms are unavailable or have limited availability due to competing priorities.  In others companies may restrict randomization when it is seen as risky for the business, for example when suboptimal treatments may lead to customer dissatisfaction.  In addition, experimental data are specific to a particular intervention and outcome, so models trained on one experiment may not generalize to other actions or contexts.  Moreover, because experiments are usually smaller in scale, the resulting models are often trained on far less data than their predictive counterparts.

Predictive models, in contrast, are abundant and versatile. A single risk score---for example, a churn probability---can be informative for multiple retention campaigns, as well as upsell offers and customer service prioritization. Likewise, a purchase-propensity score can support brand awareness, purchase-centric campaigns, and reactivation campaigns. Predictive models can be trained once, often on massive observational datasets, and then reused across many types of interventions.  However, how effective they are for any particular campaign varies.

This tension presents an opportunity: rather than treat predictive and causal models as competing alternatives, we investigate whether they can be combined effectively. Specifically, we introduce \textbf{causal post-processing (CPP)}, a family of techniques that refines proxy scores using limited experimental data. Instead of discarding predictive models, we treat their scores as a foundation and then adjust them with experimental data so they better align with causal impact.  The key question is whether CPP allows more effective causal decision making with (much) less investment in experimental data.

Put differently, CPP learns a function that transforms proxy scores so that they more closely reflect causal effects. Experiments provide the training data to learn this function, allowing a general-purpose score---such as churn risk or purchase propensity---to be tailored to a specific intervention and outcome. In this way, the same predictive model can serve as the foundation for multiple tailored variants, each aligned with a different intervention-outcome pair.

Several existing methods can be viewed through this lens, even though they have not previously been presented as causal post-processing of non-causal scores. What unifies them is the same core idea: each can be used to learn a transformation of predictive scores, guided by experimental data, so that the scores better approximate the causal effects. Framing them together as CPP highlights this shared principle, while also making room for new approaches, including one we introduce in this paper. The approaches differ mainly in how they balance flexibility and data efficiency. At one extreme, highly structured methods impose strong assumptions on the transformation, making them data-efficient but less flexible. At the other, fully flexible methods can capture complex relationships, but require more experimental data.

We examine CPP through both simulations and an empirical study. The simulations allow us to compare different approaches across a wide range of conditions, showing that no single method dominates. Instead, performance reflects an assumptions-vs-data trade-off, akin to the familiar bias--variance trade-off: methods that make stronger assumptions are more data-efficient but risk larger errors if those assumptions do not hold; methods based on weaker assumptions are more robust but require substantially more data. The empirical study, set in digital advertising, illustrates CPP in practice. It shows how different proxy scores can favor different post-processing approaches and yield different levels of improvement. Together, these analyses show when CPP adds value: it works best when proxy scores already capture a meaningful-but-imperfect causal signal, and when cost concerns limit how much experimental data is collected.

We make three main contributions. Conceptually, we establish CPP as a unifying perspective where existing and new methods can all be understood as transformations of predictive scores guided by experimental data, situated along a spectrum that trades off bias and variance. Methodologically, we introduce a new technique that lies in the middle of this spectrum, offering a practical compromise between flexibility and data efficiency. Empirically, through simulations and an application in digital advertising we show when and by how much CPP can improve intervention decisions.

\section{Scoring Models for Decision Making}\label{sec:decision-making}

Decision-making processes often rely on \textbf{scoring models}, which assign numerical values, or \textbf{scores}, to individuals or entities based on observed characteristics. These scores guide decisions in areas such as marketing, healthcare, and public policy, where they help prioritize customers, target interventions, or allocate resources.

A score typically represents an estimate of some underlying quantity relevant to a decision. For example, in customer retention, a churn score may reflect the likelihood that a customer will cancel their subscription. In healthcare, a risk score may quantify the probability that a patient will relapse. In advertising, a response score may estimate how likely a user is to purchase the product.

While these scores may be predictive of relevant behaviors, their usefulness depends on how well they align with the actual decision-making goal. In the contexts we consider, the objective is to make intervention decisions---that is, to determine whether to take an action (e.g., offering a discount, displaying an ad, or providing medical treatment) to causally influence an outcome of interest. For instance, a churn score predicts whether a customer will leave, but retention incentives are meant to prevent churn.  What matters is not just whether someone is likely to leave, but whether an intervention can (profitably) change that outcome.

This distinction is critical because decision makers often rely on predictive models that estimate outcomes---such as what would happen without any intervention---rather than true intervention effects. These models are nonetheless widely used because their predictions can serve as \textbf{proxy scores} for causal effects. For example, a customer with a low churn score will probably stay regardless of intervention, making action most likely unnecessary, while a customer with a high churn score has greater potential to be persuaded to stay. In this way, proxy scores can guide decisions, even though they do not measure causal effects directly. The challenge is that the alignment is imperfect. Some high-risk individuals may be entirely unresponsive, while some lower-risk individuals may still leave but also be persuadable.

In this work, \emph{we assume that the scoring model is given}---regardless of how it was built---\emph{and focus on the problem of refining its outputs to better support intervention decisions}. While the concepts we discuss apply to any form of scoring model, our emphasis is on predictive models: models trained on historical data to generate scores that correlate with observed behaviors or outcomes (e.g., models that estimate what would happen without intervention). The central goal is to explore whether and how the proxy scores can be refined using experimental data to better inform causal decisions.

Let \( Y^0 \) and \( Y^1 \) represent the potential outcomes for an individual under no treatment and treatment, respectively. The \textbf{individual's causal effect} is:
\[
\tau = Y^1 - Y^0.
\]
Our central concern is whether a given score \( S \) provides useful guidance for causal decisions by relating meaningfully to the individual's effect \( \tau \). Depending on the decision-making context, we may require different properties from the score. We refer to these properties as \textbf{decision-making requirements}: conditions the score must satisfy to support a specific type of decision.
\begin{itemize}
    \item If the decision requires estimating how large the effect is for each individual, the score should approximate the effect itself (\emph{effect magnitude accuracy}).
    \item If the decision requires identifying whether the effect is beneficial enough, the score should distinguish individuals whose effects exceed a critical threshold (\emph{effect classification accuracy}).
    \item If the decision involves prioritizing individuals based on those who benefit most, the score should rank individuals by their effects (\emph{effect ordering accuracy}).
\end{itemize}

\begin{figure}
\centering
\includegraphics[width=\textwidth]{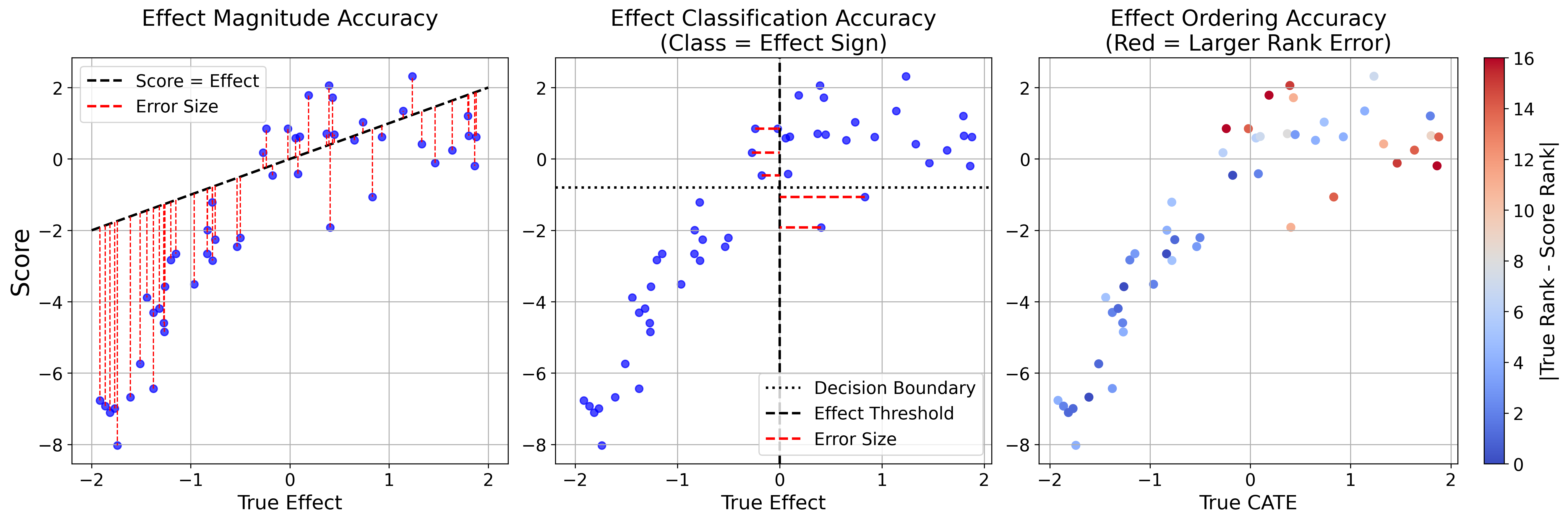}
\caption{
\textbf{Performance depends on the goal.}
Each panel visualizes a different notion of score quality using the same scores (points) plotted against the true effect. 
\textbf{Left:} For \emph{effect magnitude accuracy}, vertical errors (red lines) show the scores systematically deviate from the true effects, especially for negative effects where scores appear too large in magnitude. 
\textbf{Center:} For \emph{effect classification accuracy}, the model performs well: most predictions fall on the correct side of the decision boundary (horizontal line), and errors occur primarily for small effects near the class boundary (vertical line). 
\textbf{Right:} For \emph{effect ordering accuracy}, color indicates rank error. The scores rank individuals with negative effects well but struggle to order individuals with positive effects accurately.
}
\label{fig:model-evaluation}
\end{figure}

These requirements are conceptually distinct and correspond to different practical needs. To build intuition around them, Figure~\ref{fig:model-evaluation} provides a visual comparison. It shows the same set of proxy scores plotted against the true effect, with each panel highlighting a different requirement. We will return to this figure in each of the next three subsections to discuss each requirement in more detail.

\subsection{Effect Magnitude Accuracy}

The most stringent requirement is \textbf{effect magnitude accuracy}, the ability of scores to approximate the size of an intervention’s effect for each individual.

This requirement typically arises in settings where effect estimates are only one input to a broader decision process. For example, a clinician might weigh predicted treatment effects alongside patient-specific ethical considerations or qualitative judgments. In such cases, simply prioritizing individuals by effect size may not be enough to make effective individual decisions.

Complex resource constraints are another reason why accurate effect magnitudes may be necessary. Imagine a decision maker with a fixed budget, where individuals differ both in how costly they are to treat and in how much they would benefit. In this case, simply ranking individuals by their predicted effects may not suffice. The optimal allocation is not necessarily the set of those with the largest effects, but rather the set of individuals whose total costs fit within the budget and yield the greatest overall benefit~\citep{mcfowland2021prescriptive}.

Figure~\ref{fig:model-evaluation} (left panel), which displays the true effects as the dashed line and the scores as points, illustrates this idea. Suppose negative effects are desirable---for instance, when the intervention reduces churn risk. The vertical red lines show that if we treat the scores as effect estimates, then the effects are substantially overestimated: the score-estimates look far more negative than the effects really are. Such exaggeration is especially problematic when costs vary across individuals, because it is not enough to know that effects are negative or relatively large. Effective allocation requires accurate magnitude estimates so that benefits can be weighed against costs for each individual.

In practice, scoring models are rarely used to estimate precise causal-effect magnitudes. They are applied primarily as tools for prioritization---ranking individuals or flagging those most likely to be affected---rather than for explicit individual-level trade-off analysis. Even when scores correlate with causal effects, such as churn scores being higher for customers more likely to be influenced by a retention incentive, they are not typically interpreted as effect-size estimates, as far as we are aware.

Why, then, discuss effect magnitude at all? Because this goal arises in practical decision contexts, and a key contribution of this work is to demonstrate that non-causal scores---though not designed for this purpose---can in fact be post-processed to better reflect effect magnitudes.
Of course, no model will estimate effects perfectly, even after post-processing, so it is important to measure how closely scores approximate true effects. Let $S$ denote a score for an individual and $\tau$ their true effect. A common metric is the \textbf{Mean Squared Error (MSE)}:
\begin{equation}\label{eq:mse}
\text{MSE} = \mathbb{E}\left[\left(\tau-S\right)^2\right].
\end{equation}

\subsection{Effect Classification Accuracy}\label{sec:classification}

In many practical settings, what matters is not the exact size of the effect but whether it clears a threshold: is the intervention beneficial, or beneficial enough to warrant action? This leads to our next requirement, \textbf{effect classification accuracy}, the ability of scores to distinguish between individuals with effects above and below a critical threshold. Unlike magnitude accuracy, which focuses on precise estimates, classification accuracy is concerned with correct (binary) distinctions.

The intuition is illustrated in the center panel of Figure~\ref{fig:model-evaluation}. Two thresholds are at play. The \textbf{effect threshold} (vertical line) defines the classes: individuals with effects above the threshold (an effect above zero, in this example) belong to the positive class, and those below belong to the negative class. The \textbf{decision boundary} (horizontal line) defines how scores assign individuals to those classes---positive if their score exceeds the boundary, negative otherwise. These two thresholds need not coincide: the effect threshold reflects the intervention’s objective, while the decision boundary is chosen to optimize performance given imperfect scores. In this illustration, the effect threshold is zero, but a lower decision boundary is chosen to improve classification performance.

Most predictions in the example fall in the correct quadrants: individuals with effects above zero appear in the upper-right quadrant (true positives), while those with effects below zero appear in the lower-left quadrant (true negatives). Misclassifications appear in the off-diagonal quadrants. In this case, they are concentrated near the effect threshold, so the errors are not severe.

Effect classification accuracy is the primary concern in many real-world decision problems, where the objective is not to estimate effect magnitudes precisely but to ensure that the right individuals are targeted. In domains such as advertising, customer retention, and online recommendations, good classification accuracy often suffices, and misalignments between proxy scores and causal effects may be inconsequential as long as the model captures relative differences across individuals. Proxy scores can even outperform direct effect estimates in this role. For example, in advertising, models that predict purchase likelihood have been shown to separate high- from low-effect individuals more effectively than direct effect estimates~\citep{fernandez2022causal}.

One way to evaluate effect classification accuracy is through the \textbf{expected policy gain (EPG)} $\pi$, which measures the average net gain from treating only the estimated ``high-effect'' group. Let $\kappa$ represent the effect threshold (e.g., based on the cost of treatment), so that $\tau-\kappa$ is the net gain from treating an individual, and let $A$ denote treatment assignment based on whether an individual's score exceeds the decision boundary $\tilde\kappa$.\footnote{In the center panel of Figure~\ref{fig:model-evaluation}, the upper-left quadrant shows cases where the net gain is negative but individuals were treated, and the lower-right quadrant shows cases where the net gain is positive but individuals were not treated. These are causal misclassifications.} Then: 
\begin{align}
    \pi = \mathbb{E}[A(\tau - \kappa)], \label{eq:epg}\\
    A= \textbf{1}\{S>\tilde{\kappa}\}. \label{eq:action}
\end{align}

As shown in Appendix~\ref{App:ewm}, comparing scoring models based on $\pi$ is equivalent to comparing them based on their \textbf{effect-weighted misclassification (EWM)}:
\begin{align}\label{eq:ewm}
\text{EWM} = \mathbb{E}\left[\left|\tau-\kappa\right|\cdot\textbf{1}\left\{\tau>\kappa \neq S>\tilde{\kappa}\right\}\right], 
\end{align}
where the ``high-effect'' class is $\tau>\kappa$, the ``low-effect'' class is $\tau\leq\kappa$, and classifications are made based on $\textbf{1}\{S>\tilde{\kappa}\}$.

Classification accuracy and magnitude accuracy do not always align. A model with poor magnitude estimates may still separate individuals correctly above and below the threshold, while higher magnitude accuracy does not guarantee better classification~\citep{fernandez2022causaldecision}.

One common challenge with using scores that do not directly estimate causal effects is ensuring that the intervention provides a benefit that exceeds its cost for those who receive it---or at least that it is not harmful. This issue can often be addressed by adjusting the decision boundary. Specifically, Equation~\eqref{eq:ewm} shows how treatment assignment depends on the decision boundary $\tilde{\kappa}$, which can be calibrated to balance costs and expected benefits.

This calibration is simpler in resource-constrained settings, where only a limited number of individuals can be treated due to budget, time, or operational limits. For example, if only 1,000 individuals can be treated, the decision boundary $\tilde{\kappa}$ can be set to the score of the 1,000th highest-ranked individual. In that case, whether it would also benefit the 2,000th or 3,000th individual is immaterial---only the top 1,000 can be treated. As long as scores effectively rank individuals by their effects and the intervention has a positive net benefit on the cutoff individual, precise magnitude estimation is unnecessary.

The broader point is that many applications that seem to require accurate effect magnitudes in fact only require accurate effect classification. As long as scores reliably separate individuals above and below the relevant threshold, they can support good decisions. There are, of course, cases where ranking alone is insufficient and more precise estimates are needed (e.g., budget constraints with variable costs, as discussed above). Ultimately, the goal is not to minimize estimation error, but to improve decision making.

\subsection{Effect Ordering Accuracy}

While classification accuracy evaluates decisions at a specific threshold, we may need greater flexibility. Budgets may change, new resources may become available, or constraints may tighten. For example, a company might shift from targeting the top 1,000 customers to only the top 500. In such settings, rather than assigning individuals to fixed groups, the goal is to establish a ranking where those with higher scores are more likely to benefit from the intervention.

We refer to this requirement as \textbf{effect ordering accuracy}, the ability of scores to rank individuals by the magnitude of their causal effects. If ordering accuracy holds, then as scores increase, expected intervention effects should tend to increase as well. 

Ordering accuracy is naturally assessed through ranking-based metrics. We adopt the rank-weighted average treatment effect (RATE) framework~\citep{yadlowsky2024evaluating}, which generalizes the evaluation of treatment prioritization rules. The core building block is the Targeting Operator Characteristic (TOC), which measures the effect of treating the top-ranked $q$ fraction of individuals relative to treating at random:
\begin{equation}\label{eq:toc}
    \text{TOC}(q) = \mathbb{E}\left[\tau|F(S)\geq 1-q\right] - \mathbb{E}[\tau],
\end{equation}
where $F(\cdot)$ is the cumulative distribution function (CDF) of the scores. The RATE metric aggregates TOC values across all $q$ using a weighting function $\alpha(q)$:
\begin{equation}\label{eq:rate}
    \text{RATE} = \int_0^1 \alpha(q)\cdot \text{TOC}(q)~dq.
\end{equation}

Different choices of $\alpha(q)$ emphasize different ranking priorities. Two common options are a uniform weight, $\alpha(q)=1$, known as \textbf{AUTOC (Area under the TOC curve)}, and a linear weight, $\alpha(q)=q$, known as the \textbf{Qini}. AUTOC weights all quantiles equally, giving proportionally more influence to the top of the ranking, while the Qini increases the weight with $q$, emphasizing overall ranking quality across the population.

Importantly, strong performance on magnitude or classification accuracy does not imply strong ordering accuracy, and vice versa. The right panel of Figure~\ref{fig:model-evaluation} illustrates this point. Each point is colored by ranking error. Scores rank individuals with negative effects well (mostly blue), even though the left panel shows large magnitude errors for those same individuals. By contrast, individuals with positive effects are ranked poorly (mostly red), despite having more accurate magnitude estimates. At the same time, the center panel shows that classification performance remains strong overall---large magnitude or ranking errors do not necessarily translate into large classification errors.

The contrast shows that magnitude, classification, and ordering accuracy are distinct. Scores may estimate effect sizes poorly yet rank individuals well, or classify effectively at a particular threshold while failing to preserve the correct overall order. These differences can materially affect decisions.

The distinction mirrors conventional predictive modeling. In classification tasks, we evaluate probability estimation accuracy, threshold-based classification accuracy, and ranking metrics like AUC separately, because each captures a different dimension of model quality. The same logic applies here: effect magnitude, classification, and ordering accuracy reflect different goals, and gains in one do not guarantee gains in another.

\section{Causal Post-Processing}\label{sec:post-processing}

Causal Post-Processing is a family of methods that use experimental data to refine proxy scores, aligning them more closely with causal effects for better decision making. As discussed earlier, organizations often deploy predictive models trained on large-scale observational data that predict an outcome of interest, such as the probability of churn or purchasing. The resulting \textbf{base scores} $\tilde{S}$ are then used to guide targeting decisions, not because they are causal estimates, but because they are expected to correlate with treatment effects. CPP leverages experimental data to learn a \textbf{post-processor} $\Psi$ that maps $\tilde{S}$ (and, optionally, covariates $X$) into a post-processed score $S$:
\begin{equation}
    S = \Psi(\tilde{S}, X).
\end{equation}

The goal is for $S$ to align more closely with causal effects, so that it better supports the decision-making requirements introduced in Section~\ref{sec:decision-making} (for effect magnitude estimation, effect classification, or effect-based ordering).

Formally, we assume access to experimental data from a randomized trial on a random sample of the target population.\footnote{These data could be, for example, from an A/B test.  Note that A/B tests are usually powered for conclusions about ATEs, and thus the data collected will likely be small compared to what would be needed to learn a full-blown CATE model of individual effects.} 
The data for this experimental sample comprise:
\begin{itemize}
\item \textbf{Treatment assignment}: $T \in \{0,1\}$, indicating whether the individual received the intervention ($T=1$) or not ($T=0$).
\item \textbf{Outcome of interest}: $Y \in \mathbb{R}$, the observed response (e.g., purchase, churn).
\item \textbf{Base score}: $\tilde{S}$, the unprocessed score from the pre-existing model.
\item \textbf{Pre-treatment covariates} (if available): $X$, pre-treatment features that may help refine $\tilde{S}$.
\end{itemize}

We make three main assumptions. First, treatment assignment in the experiment is ignorable due to randomization, so causal effects can be estimated without confounding even without conditioning on the base score or covariates: $Y(0), Y(1) \perp T$. Second, we assume generalizability: the experimental sample represents the target population of interest. We can formalize this as $Y(1)-Y(0) \perp R$, where $R=1$ denotes the experimental sample and $R=0$ denotes the target population. Third, we assume SUTVA \citep{rubin1990comment}, meaning that there is no interference between units and no hidden versions of the treatment.

The optional covariates $X$ are assumed to be available in the target population where post-processed scores $S$ will be applied, but they may or may not overlap with those used to estimate $\tilde{S}$. They may coincide when the covariates are stable and either logged consistently or straightforward to compute.  However, additional covariates may be collected when running the experiments.  Furthermore, the covariates used to estimate $\tilde{S}$ may not be available.  For example, systems delivering digital ads based on massive fine-grained, dynamic features may not log those features for online individuals at arbitrary points in time, and the feature-generation system used by the ad-serving decision process may not be accessible to the experimental platform~\citep{gordon2023close}.

As mentioned, our task is to learn a post-processor $\Psi$ such that $S = \Psi(\tilde{S}, X)$ better satisfies the relevant decision-making requirement. We view this as a general paradigm under which several existing techniques can be reinterpreted as forms of CPP. We group these methods into three categories, which we discuss in detail in the following subsections. 
The key distinction across the categories is how strongly they assume that the base score is structurally aligned with treatment effects. \textbf{Calibration post-processing} makes the strongest assumptions; in our case, we examine a method that assumes a linear relationship between the base score and treatment effects. \textbf{Residual post-processing} relaxes assumptions by allowing covariates to explain additional variation. \textbf{Covariate post-processing} makes the weakest assumptions by treating the score as just another covariate in a general causal model. 

Critically, the value of distinguishing these three approaches is that none is universally ``best.'' Their performance depends on two factors: (i) the extent to which their structural alignment assumptions hold in practice, and (ii) the amount of experimental data available to support estimation. Methods that make strong assumptions (e.g., calibration post-processing) can be highly data-efficient, but may work poorly if those assumptions are violated. Methods that make weak assumptions (e.g., covariate post-processing) are more robust to misspecification, but require substantially more data to perform well. Residual post-processing strikes a middle ground by leveraging structure in the score while allowing additional flexibility through covariates.  It is important of course to keep in mind that performance is relative: with a small amount of data, methods like calibration post-processing may still perform best despite violations of their assumptions, just as high-bias methods like linear regression often outperform more flexible methods for predictive modeling when data is scarce.

As Figure~\ref{fig:trade-off-analysis} illustrates, the trade-off is straightforward: stronger assumptions reduce data requirements but increase vulnerability to misspecification, while methods that make weaker assumptions are more robust but require substantially more data.

\begin{figure}
    \centering
    \begin{tikzpicture}[
        scale=1,
        every node/.style={font=\sffamily\small},
    method/.style={draw, ellipse, thick, 
                   minimum width=3.8cm, minimum height=1.6cm,
                   align=center}
    ]
        \draw[->, very thick] (-1.2,0) -- (8.8,0);
        \draw[->, very thick] (-1,-0.2) -- (-1,5.0);

        \node[font=\sffamily\bfseries\large, anchor=center] at (4,-1) {Assumptions};
        \node[font=\sffamily\bfseries\large, align=center] at (-3,2.7) {Data\\Requirements};

        \node[anchor=north] at (0.2,-0.2) {Weak};
        \node[anchor=north] at (8,-0.2) {Strong};
        \node[anchor=east] at (-1.3,0.5) {Low};
        \node[anchor=east] at (-1.3,4.7) {High};

\coordinate (COV) at (1.3, 4.2);
\coordinate (RES) at (4.0, 2.7);
\coordinate (CAL) at (6.7, 1.2);

\node[method, fill=blue!10]    at (COV) {\textbf{Covariate}\\post-processing};
\node[method, fill=blue!10] at (RES) {\textbf{Residual}\\post-processing};
\node[method, fill=blue!10]   at (CAL) {\textbf{Calibration}\\post-processing};

    \end{tikzpicture}
    \caption{\textbf{Assumptions--Data trade-off in CPP.} Methods differ in how strongly they assume that the base score is structurally aligned with treatment effects. Calibration post-processing makes the strongest assumptions and is thus the most data-efficient, but it performs poorly if those assumptions are violated. Covariate post-processing makes the weakest assumptions and is more robust to misspecification, but it requires substantially more data. Residual post-processing lies in between, leveraging the score while allowing additional variation to be explained by covariates.}
    \label{fig:trade-off-analysis}
\end{figure}

\subsection{Calibration Post-Processing} \label{sec:calibration-pp}

Scores are often used to rank individuals, which implicitly assumes that treatment effects increase monotonically with the score. \textbf{Calibration post-processing} makes this assumption explicit: it applies a monotonic transformation to align scores more closely with causal effects. The idea is analogous to probability calibration in predictive modeling, where methods like Platt scaling or isotonic regression adjust uncalibrated classifiers (e.g., SVMs) to produce well-calibrated probabilities.

As we discussed earlier, models can be good at detecting effect heterogeneity while still having poor effect magnitude accuracy. Ideally, if magnitude accuracy is a concern, a model that reports an effect of $t$ for a stratum of individuals should satisfy the calibration property that---on average---those individuals have an effect close to $t$:  
\begin{equation}
\mathbb{E}[\tau \mid \hat\tau = t]\approx t.    
\end{equation}

In practice, machine learning estimators of heterogeneous effects can violate this property, even when they rank individuals well~\citep{leng2024calibration}. This is the causal analogue of a classifier with high ROC AUC but poorly calibrated probabilities.

Effect calibration targets this issue. To date, however, it has been considered mainly in the context of causal-effect models, where the goal is to take the estimated effects and improve their calibration. Yet there is no reason why the same idea could not be applied more broadly. In particular, it can also be used to refine proxy scores that correlate with treatment effects but were never designed to estimate them directly. Concretely, instead of calibrating $\hat\tau$ from a causal model, we calibrate the base score $\tilde{S}$ itself, learning a transformation $\Psi_{\text{cal}}$ so that the post-processed score $=\Psi_{\text{cal}}(\tilde{S})$ better matches (conditional) average effects estimated with experimental data. 

In this study, we focus on the approach to effect calibration introduced by \citet{leng2024calibration}. Their method uses experimental data to learn a simple linear rescaling of model outputs so that predicted effects better match observed averages. While the mapping $\Psi_{\text{cal}}$ could in principle be nonparametric (e.g., isotonic regression), we emphasize the linear case here because it is straightforward to implement and performs well with limited data.

The method proceeds in two steps. First, given experimental data---including treatment assignment, observed outcomes, and base scores---individuals are grouped into percentile bins based on their scores (e.g., bottom 10\%, 10th--20th percentile, etc.).\footnote{The number of bins can be chosen as a hyperparameter and tuned using out-of-sample data.} For each bin $\mathcal{B}$, two quantities are computed: the average treatment effect $\tau_{\mathcal{B}}$, estimated as the difference between mean outcomes of treated and untreated individuals, and the mean base score $\tilde{S}_{\mathcal{B}}$.  

Second, a simple linear regression is fit to map average base scores to average effects:
\begin{equation}\label{eq:bin-regression}
    \mathbb{E}[\tau_\mathcal{B}\mid\tilde{S}_\mathcal{B}] \approx \alpha_0 + \alpha_1 \cdot \tilde{S}_\mathcal{B}.
\end{equation}

The estimated parameters $\hat\alpha_0, \hat\alpha_1$ define the calibration function, which can then be applied to any base score at the individual level:
\begin{equation}\label{eq:linear-calibration}
    \Psi_{\text{cal}}(\tilde{S}) =\hat\alpha_0 + \hat\alpha_1 \cdot \tilde{S}.
\end{equation}

This linear transformation preserves the original ranking while rescaling scores to better reflect actual treatment effects. Similar to conventional probability calibration techniques in predictive modeling (e.g., Platt scaling), effect calibration leaves the ordering unchanged but adjusts the scale so scores can be interpreted as effect estimates.

Calibration is most useful for improving effect magnitude accuracy, as it aligns scores with empirical estimates of average effects. It can also enhance classification accuracy when intervention decisions hinge on ensuring that benefits exceed costs: by shifting the score scale, calibration helps place the decision boundary in the right location, reducing the risk of treating individuals for whom the intervention is ineffective or harmful. However, calibration does not alter the ordering of individuals, so it cannot improve effect ordering accuracy, nor does it improve effect classification accuracy in settings where decisions depend only on targeting the top individuals with largest scores.

This limitation is the cost of making strong structural assumptions. However, the key advantage is data efficiency. For instance, the calibration method we examine here requires learning only two parameters from experimental data. The approaches we discuss next relax these assumptions and address a broader range of decision-making requirements, but they also typically demand more experimental data to perform well.

\subsection{Covariate Post-Processing}

At the other extreme from calibration, \textbf{covariate post-processing} avoids imposing a structural assumption (such as a monotonic or parametric link) between proxy scores and causal effects. The base score $\tilde{S}$ is treated simply as a covariate in a causal model, possibly alongside many others, and relies on machine learning to uncover how treatment effects vary with $\tilde{S}$. The most direct implementation, and the one we focus on in this study, is to estimate the \textbf{Conditional Average Treatment Effect (CATE)} conditional on $\tilde{S}$ (and possibly other features $X$\footnote{As discussed above, the additional covariates $X$ may or may not be the covariates used to estimate $\tilde{S}$.  Additional covariates may be available in the experimental data, and original covariates may not be available.}):
\begin{equation}
    \Psi_{\text{cov}}(\tilde{S}, X) = \hat{\mathbb{E}}[\tau \mid \tilde{S}, X].
\end{equation}
This implementation directly targets effect magnitude accuracy, aligning scores with individual-level treatment effects. If the decision objective were different, we could change the learning target accordingly. For instance, if the goal were effect ordering, one could train the model to maximize rank correlation between predicted and true effects \citep{devriendt2020learning}. If the goal were classification into high- and low-effect groups, the problem could instead be framed as a weighted classification task in which the target corresponds to the treatment with the largest potential outcome, and weights reflect the relative importance of correct classification \citep{zhang2012estimating}. In this sense, covariate post-processing provides a flexible framework that can support any of the decision-making requirements discussed in Section~\ref{sec:decision-making}.

Covariate post-processing is the most general but also the most data hungry of the approaches we discuss. Unlike calibration post-processing, which assumes that the score's ranking already reflects the ordering of treatment effects, covariate post-processing makes no such assumption. It allows the model to learn whatever relationship the data supports---linear or nonlinear, monotone or not. In principle, any method for heterogeneous treatment effect estimation can be used here~\citep[e.g.,][]{Hill2011Bayesian, nie2021quasi}. The trade-off is that the algorithm must learn the full relationship between scores, features, and outcomes from experimental data, which can lead to underperformance when data are limited~\citep{fernandez2022causal, fernandez2025observational}. 

Covariate post-processing can also be framed as a form of feature engineering for causal estimation. Instead of relying on the raw inputs that generated the base score $\tilde{S}$, it treats $\tilde{S}$ itself as a summary statistic that may capture meaningful variation in treatment effects across individuals. This idea parallels other scalar scores in causal inference, such as the propensity score~\citep{rosenbaum1983central} or prognostic score~\citep{hansen2008prognostic}, which condense complex feature sets into a single number useful for estimation---though their purpose is typically adjustment for confounding rather than modeling heterogeneity. Here, the base score is viewed as an (imperfect) proxy for treatment-effect heterogeneity. Even if it is not itself a causal estimate, incorporating it into the causal model can improve statistical efficiency, particularly when experimental data are limited.

This concept was explored by \citet{peysakhovich2016combining} for integrating confounded observational effect estimates with causal model learning.  Specifically, they showed that biased observational effect estimates can serve as valuable input features for causal models trained on experimental data. Despite their bias, these estimates often remain predictive of effects---for example, larger observational estimates may still align with larger causal effects, even in the presence of confounding. 

The same logic applies to proxy scores. Although $\tilde{S}$ may not directly estimate effects, it may still carry information about how those effects vary across individuals, enabling more accurate estimation. The challenge is that simply including $\tilde{S}$ as an undifferentiated covariate may prevent the model from capturing this relationship effectively. When proxy scores are considered for decision making, it is usually because there is a strong belief that they capture meaningful variation in treatment effects. This suggests that proxy scores may be better exploited when modeling strategies consider their presumed link to treatment effects explicitly, rather than treating them as generic covariates. 

A clear example comes from \citet{athey2025machine}, who showed that predictions of baseline outcomes ($\mathbb{E}[Y^0 \mid X]$) can improve effect estimation once their relationship to treatment effects is explicitly accounted for. This was not cast as post-processing, but the lesson applies here: when proxy scores have a well-defined meaning---such as baseline predictions---and we acknowledge their causal role (e.g., individuals with higher baselines tending to have larger effects), this can help guide the covariate post-processing to improve intervention decisions.

Of course, proxy scores rarely capture the entire causal signal. They may align with treatment effects in systematic ways but still leave important residual variation unexplained. This motivates an intermediate strategy: calibrate the main signal carried by the score, then explicitly model what remains---a strategy we develop in the next section.

\subsection{Residual Post-Processing}\label{section:residual-pp}

Residual post-processing takes a middle ground between calibration and covariate post-processing. It begins from the structural link imposed by calibration but adds a residual component to capture systematic deviations from that link. In other words, calibration assumes the score structure is correct, while covariate post-processing assumes no particular relationship between $\tilde S$ and treatment effects. Residual post-processing assumes the score provides a useful starting point and then corrects what calibration misses.

Formally, we decompose the post-processing into two components:
\begin{equation}\label{eq:residual}
\Psi_{\text{res}}(\tilde{S}, X) = \Psi_{\text{cal}}(\tilde{S}) + \delta(X),
\end{equation}
where $\Psi_{\text{cal}}(\tilde{S})$ is a calibrated transformation of the base score and $\delta(X)$ models residual heterogeneity explained by covariates.\footnote{This formulation assumes additivity between $\tilde S$ and $X$. In practice, one could also include $\tilde S$ as an input in $\delta$ to allow interactions between $\tilde S$ and $X$.} In this sense, residual post-processing extends calibration with a second stage focused on residual errors.

The key advantage of residual post-processing is that it balances data efficiency with robustness. If $\tilde{S}$ is highly informative, the calibrated component $\Psi_{\text{cal}}(\tilde{S})$ explains most of the variation and the residual model contributes little. If $\tilde{S}$ is weak, the residual term $\delta(X)$ carries more weight, possibly approaching a fully general causal model. In between, the residual stage can correct systematic distortions left by calibration.

Crucially, residual post-processing is not meant to universally outperform calibration or covariate post-processing. As illustrated in Figure~\ref{fig:trade-off-analysis}, each method reflects a different point on the trade-off spectrum: calibration relies most on the score but is highly data-efficient; covariate post-processing relies least on the score but requires more data, and residual post-processing sits in the middle, providing a practical compromise.

One simple way to implement residual post-processing is with a \textbf{causal residual tree}. The idea is to let the calibrated score, $\Psi_{\text{cal}}(\tilde{S})$, capture the main signal from the proxy score, and then fit a tree-structured model to the remaining variation, $\delta(X)$. Other machine learning methods could also be used for modeling $\delta(X)$, but we use trees here to illustrate the approach concretely.

Estimation proceeds leaf by leaf. For each segment $\Omega$ (a leaf) of the partitioned feature space, the model assigns a constant residual correction $\delta$ chosen so that the corrected score, $\Psi_{\text{cal}}(\tilde{S})+\delta$, best matches the average treatment effect in that leaf. Concretely, $\delta$ is chosen to minimize 
\begin{align}\label{eq:bias-mse}
\text{MSE}\left(\delta ; \Omega\right) =\mathbb{E}\left[(\tau-\Psi_{\text{cal}}(\tilde{S})-\delta)^2\mid X\in \Omega\right].
\end{align}

The minimizer is the average bias of the calibrated score relative to the leaf's average effect:
\begin{equation}\label{eq:mean_bias}
    \delta^*(\Omega)=\mathbb{E}\left[\tau\mid X\in \Omega\right] - \mathbb{E}\left[\Psi_{\text{cal}}(\tilde{S})\mid X\in \Omega\right].
\end{equation}

So, given a sample $\mathcal{D}$ drawn from $\Omega$, we estimate $\delta^*$ as the difference between the leaf's empirical treatment effect and the mean calibrated score:
\begin{align}
\hat\delta(\mathcal{D}) &= (\hat\mu_y(1)-\hat\mu_y(0))-\hat\mu_b, \label{eq:ee_estimate}\\
\hat\mu_y(t) &=\frac{1}{N(t)}\sum_{i\in \mathcal{D}:t_i=t} y_i, \nonumber\\
\hat\mu_b &= \frac{1}{N(1)+N(0)}\sum_{i\in \mathcal{D}}\Psi_{\text{cal}}(\tilde{S}_i). \nonumber
\end{align}
Each observation $i \in \mathcal{D}$ has a treatment assignment $t_i$, outcome $y_i$, and calibrated score $\Psi_{\text{cal}}(\tilde{S}_i)$. $N(t)$ is the number of treated ($t=1$) or control ($t=0$) units.

To construct the leaves, the residual algorithm uses tree induction to recursively partition the data based on its features. At each step, it evaluates candidate splits of the feature space by comparing how much they reduce MSE. Formally, let $\Omega_P$ denote the parent node with sample $\mathcal{D}_P$. A candidate split $\ell$ divides it into two child nodes, $\Omega_L^\ell$ and $\Omega_R^\ell$, with corresponding subsamples $\mathcal{D}_L^\ell$ and $\mathcal{D}_R^\ell$. The improvement from split $\ell$ is estimated as:
\begin{align}\label{crite_ee_estimate}
    \widehat{\text{MSE}}\left(\hat\delta(\mathcal{D}_P)\right)-\frac{|\mathcal{D}_L^\ell|}{|\mathcal{D}_P|}\cdot \widehat{\text{MSE}}\left(\hat\delta(\mathcal{D}_L^\ell)\right)-\frac{|\mathcal{D}_R^\ell|}{|\mathcal{D}_P|}\cdot \widehat{\text{MSE}}\left(\hat\delta(\mathcal{D}_R^\ell)\right).
\end{align}
where $|\mathcal{D}|$ represents the size of sample $\mathcal{D}$. The algorithm selects the candidate split that maximizes this criterion.

The main challenge is that MSE cannot be estimated directly because it depends on the unobserved effect $\tau$. To overcome this, we follow a similar approach to the one introduced in~\cite{athey2016recursive} and aim to minimize a modified version of the MSE. As in~\cite{athey2016recursive}, the modification does not affect how candidate splits are ranked. Here is the modification:
\begin{align}
\widetilde{\text{MSE}}\left(\delta;\Omega\right)&= \mathbb{E}\left[(\tau-\Psi_{\text{cal}}(\tilde{S})-\delta)^2 -(\tau-\Psi_{\text{cal}}(\tilde{S}))^2|X\in\Omega\right]\\
&=\delta^2 -2\delta\cdot\delta^*(\Omega).
\end{align}

As mentioned, we estimate $\delta^*$ with $\hat\delta$, leading to the following empirical estimate of $\widetilde{\text{MSE}}$:
\begin{equation}
\widehat{\text{MSE}}\left(\hat\delta\right)=-\hat\delta^2.
\end{equation}

In the empirical analyses that follow, we extend the causal residual tree into a forest by averaging across multiple trees with bagging.

\section{Simulation Study}

The purpose of our simulation study is to understand the circumstances under which the different types of CPP succeed or fail. An empirical study would only tell us that a method works (or fails) in a particular dataset; even across multiple datasets, we could not directly manipulate the conditions to uncover \textit{why}. Simulations allow us to do exactly that: by constructing controlled environments, we can vary the underlying factors systematically and isolate their effects.

Specifically, we examine two critical dimensions that map directly to the trade-off in Figure~\ref{fig:trade-off-analysis}. First, we vary the correlation between the base score and the true treatment effect, which captures the extent to which structural alignment holds. When correlation is high, methods that lean heavily on the base score are well supported; when it is low, such methods risk severe misspecification. Second, we vary the amount of experimental data available, which determines how difficult it is to extract signal and how reliable the resulting estimates are.

We focus on the common scenario where base scores are predictions of baseline outcomes---that is, outcomes in the absence of intervention. This choice is motivated by three considerations. 

First, baseline outcome predictions are widely used in practice across domains such as customer retention~\citep{ascarza2018pursuit}, targeted advertising~\citep{stitelman2011estimating}, nudging~\citep{athey2025machine}, precision medicine~\citep{kent2020predictive}, and recommender systems~\citep{gao2024causal}.

Second, baseline outcome estimates have broad potential because they capture what would happen without intervention. Unlike causal models, which are typically tailored to a specific intervention-outcome pair, baseline predictions can inform a wide range of interventions aimed at shifting that outcome. This generality makes them especially appealing as a foundation for CPP, which can adapt them to different decision-making requirements in various settings.

Third, baseline outcome predictions are particularly valuable when experimental data is limited. Because they reflect what is observed in the absence of intervention, they provide a practical starting point for decision making when accurate estimation of individual-level causal effects is infeasible.\footnote{Estimating individual-level causal effects often requires very large samples, and what qualifies as ``sufficiently large'' depends on the context. In some applications, even tens of millions of observations may be insufficient~\citep{fernandez2023comparison}.} Such situations are common---for example, with new interventions (like a novel retention incentive or a new ad campaign) or when interventions carry large risks, as in medicine.  In these cases, baseline predictions often become the default basis for action, and CPP offers a principled way to strengthen them by integrating whatever experimental evidence is available.

\subsection{Data-Generating Process}

As just discussed, our design varies two levers that map to the trade-off in Figure~\ref{fig:trade-off-analysis}:
(i) the alignment between the base score and the true treatment effect, and
(ii) the amount of experimental data available (sample size), which governs how precisely effect heterogeneity can be learned.

\paragraph{Lever 1: Alignment (correlation $\rho$).}
We construct two latent components and mix them to obtain effects whose correlation with the base score is exactly $\rho$.
Let $f(X)$ be the latent component that will drive the baseline outcome (and hence the base score), and let $g(X)$ be a component uncorrelated to $f(X)$. After standardizing
$f$ and $g$ to zero mean and unit variance, we define
\begin{equation}
\tau_{\text{std}} \;=\; \rho\, f_{\text{std}}(X) \;+\; \sqrt{1-\rho^2}\, g_{\text{std}}(X),    
\end{equation}
where the subscript ``std'' denotes standardization to zero mean and unit variance.

We then scale and shift to obtain the effect
\begin{equation}
    \tau \;=\; \mu_{\Delta} \;+\; \sigma_{\Delta}\,\tau_{\text{std}},    
\end{equation}
 where $\mu_{\Delta}$ sets the average treatment effect (ATE), and $\sigma_{\Delta}$ controls the degree of heterogeneity in treatment effects. 

We set the expected baseline outcome as
\begin{equation}
\theta \;=\; \mathbb{E}[Y^0\mid X] \;=\; \mu_0 \;+\; \sigma_0\, f_{\text{std}}(X),
\end{equation}
where $\mu_0$ sets the average baseline outcome, and $\sigma_0$ controls the dispersion explained by $X$.

Correlation is invariant to affine transformations, so $\mathrm{Corr}(\theta,\tau)=\rho$ by construction.

\paragraph{Lever 2: Data availability (sample size $n$).}
Observed outcomes include idiosyncratic noise,
\begin{equation}
Y \;=\; \theta \;+\; T\cdot \tau \;+\; \epsilon,\qquad \epsilon \sim \mathcal{N}(0,\sigma_\epsilon^2),
\end{equation}
where T is the treatment indicator variable, and $\sigma_\epsilon^2$ is held constant in the main experiments. We vary the experimental training size $n$, which reduces estimation error (via standard $\sqrt{n}$-type concentration) without changing the underlying signal or noise. Larger $n$ thus increases the precision with which effect heterogeneity can be learned. Appendix~\ref{app:extended} shows that reducing $\sigma_\epsilon^2$ produces an analogous improvement in estimation.

\paragraph{Experiment and base scores.}
We assume a properly conducted randomized experiment,
\begin{equation}
T \sim \mathrm{Bern}(0.5), \qquad T \perp (X, Y^0, Y^1).    
\end{equation}

As mentioned before, we focus on the case where the base scores estimate baseline outcomes. To isolate the roles of alignment and noise, we employ an oracle baseline score,
\begin{equation}\label{eq:base_scores_sim}
\tilde{S} \;=\; \mathbb{E}[Y^0\mid X] \;=\; \theta,    
\end{equation}
but estimation error in $\tilde{S}$ could be added without affecting how $\rho$ and $n$ are used to manipulate alignment and data availability.

In this setup, we assume that the covariates \(X\) used to compute the base scores are the same features available in both the experimental sample and the target population. The simulations could easily be modified to relax this assumption---for example, if the system generating \(\tilde{S}\) relies on features not logged or accessible in the experimental setting. In such cases, proxy scores become a stronger baseline because they encode information unavailable in the experimental data.

\paragraph{A simple, interchangeable feature model.}
The specific relationship between $(f,g)$ and the observed features $X$ does not affect the two levers above. For simplicity, we use $r$ independent binary features and linear relationships:
\begin{equation}
X_j \sim \mathrm{Bern}(0.5)\ \text{ i.i.d.},\quad
f(X)=\sum_{j=1}^r \alpha_j X_j,\ \ \ g_0(X)=\sum_{j=1}^r \omega_j X_j,\quad
\alpha_j,\omega_j \overset{\text{i.i.d.}}{\sim}\mathcal{N}(0,1).    
\end{equation}
We obtain $g(X)$ orthogonal to $f(X)$ via the Gram–Schmidt process:
\begin{equation}
g(X) \;=\; g_0(X) \;-\; \gamma\, f(X),\qquad
\gamma \;=\; \frac{\mathrm{Cov}(f(X),g_0(X))}{\mathrm{Var}(f(X))}.    
\end{equation}
We then standardize $f$ and $g$ before forming $\tau_{\text{std}}$ as discussed above.
This linear/i.i.d. choice is purely for simplicity---nonlinearities, interactions, or correlated $X$ can be substituted without altering control over the two key dimensions (alignment $\rho$ and data availability $n$).

\subsection{Experimental Setup}\label{sec:exp_setup}

We benchmark methods across a grid of scenarios that vary the two levers:  
\begin{itemize}
\item \textbf{Score–effect alignment.} $\rho \in \{0.1, 0.3, 0.5, 0.7, 0.9\}$, from weak to high alignment.  
\item \textbf{Sample size.} $n \in \{250, 500, 1000, \ldots, 64{,}000\}$ experimental observations.  
\end{itemize}

We fix the remaining parameters at $r=20$, $\mu_0=0$, $\sigma_0^2=1$, $\mu_\Delta=0.1$, $\sigma_\Delta^2=0.1$, and $\sigma_\epsilon^2=1.6$. Results are qualitatively similar under other settings. The location parameters mainly affect classification: $\mu_0$ shifts scores relative to the decision boundary, and $\mu_\Delta$ controls the share of individuals who benefit from treatment. We set $\sigma_0^2=1$ to define the scale. Parameters $r$, $\sigma_\epsilon^2$, and $\sigma_\Delta^2$ also affect the difficulty of estimating effect heterogeneity, but in our main experiments we vary sample size to control it. Other levers---such as increasing noise (Appendix~\ref{app:extended}), adding features to raise functional complexity, or reducing $\sigma_\Delta$ to weaken the signal---are analogous ways to make estimation harder.

We benchmark the following approaches for supporting intervention decisions:
\begin{itemize}
    \item \textbf{Average treatment effect (ATE).} Predict each individual's effect as the ATE ($\mu_{\Delta}=0.1$); a simple baseline for effect magnitude accuracy.
    \item \textbf{Base score only (BS).} Use the base score $\tilde{S}$ from Equation~\eqref{eq:base_scores_sim} without any causal post-processing (requires no experimental data).
    \item \textbf{Causal modeling (CM).} Causal forest \citep{wager2018estimation} that \emph{does not} include $\tilde{S}$ (estimates effects from $X$ alone).
    \item \textbf{Calibration post-processing (P--CAL).} Calibration of $\tilde{S}$ using the approach proposed by~\citet{leng2024calibration}.
    \item \textbf{Residual post-processing (P--RES).} The method introduced above: apply calibration post-processing to $\tilde{S}$, then model residuals with a forest that ensembles causal residual trees.
    \item \textbf{Covariate post-processing (P--COV).} Causal forest using $\{\tilde{S},X\}$ as features (same learner as CM but augmented with the base score).
\end{itemize}

Causal forests are used for CM and P--COV because they are a standard, well-studied approach to heterogeneous effect estimation and, like P--RES, are tree-based. Using tree-structured learners across the board keeps the comparisons of different post-processing techniques on even footing, to help isolate differences due to the type of CPP.

We benchmark the methods using the decision-making requirements from Section~\ref{sec:decision-making}: MSE for effect magnitude accuracy, EPG for effect classification accuracy, and AUTOC and Qini for effect ordering accuracy. For classification, we set the thresholds at $\kappa=\tilde{\kappa}=0$, which corresponds to aiming to target all individuals who benefit from treatment.\footnote{When targeting decisions are based on the unprocessed base scores (BS), this is roughly equivalent to treating the top 50\% of individuals. In our simulations, however, the average treatment effect is set to be positive, so a larger proportion of individuals benefit and should be targeted.}

Each reported point averages 100 independent simulations. In every simulation we:
(i) sample dataset-level coefficients $\{\alpha_j,\omega_j\}$,
(ii) generate an experimental dataset of 64{,}000 observations,
(iii) create training subsets of size $n\in\{250,500,1000,\ldots,64000\}$ for all methods that use experimental data (BS uses none),
and (iv) evaluate on a separate test set of 50{,}000 observations.

\subsection{Which Methods Work When?}

The four heatmaps in Figure~\ref{fig:sim-heatmap} show, for each $(\rho,n)$ pair, the method that performed best under each decision requirement (magnitude, classification, ordering). The results mirror the conceptual trade-off illustrated earlier in Figure~\ref{fig:trade-off-analysis}. A very similar pattern is observed when we vary estimation difficulty by varying idiosyncratic noise instead of sample size (Appendix~\ref{app:extended}).

\pgfplotstableread{./simulation_figures/heatmap_noise_fixed.dat}{\heatmap}

\begin{figure}
\centering
\begin{subfigure}{0.45\textwidth}
    \subcaption{MSE}\label{subfigure:sim-heatmap-mse}
    \begin{tikzpicture}[trim axis left]
        \begin{axis}[HeatmapNoiseFixed, xlabel={}] 
          \addplot[
            matrix plot*, 
            mesh/rows=5,
            mesh/cols=9,
            point meta=explicit
          ] table[x=x, y=y, meta=cat_mse] {\heatmap};
          \node at (axis cs:0.5, 1) {ATE};
          \node at (axis cs:5, 0) {CM};
          \node at (axis cs:2, 4) {P-CAL};
          \node at (axis cs:4, 3) {P-RES};
          \node at (axis cs:6.5, 2) {P-COV};
          \node[font=\scriptsize, align=center] at (axis cs:2, 1) {P-\\COV};
          \node[font=\scriptsize, align=center] at (axis cs:3, 2) {P-\\COV};
        \end{axis}
    \end{tikzpicture}
\end{subfigure}
\hfill
\begin{subfigure}{0.45\textwidth}
    \subcaption{EPG}\label{subfigure:sim-heatmap-epg}
    \begin{tikzpicture}[trim axis left]
    \begin{axis}[HeatmapNoiseFixed, xlabel={},ylabel={}] 
      \addplot[
        matrix plot*, 
        mesh/rows=5,
        mesh/cols=9,
        point meta=explicit
      ] table[x=x, y=y, meta=cat_epg] {\heatmap};
      \node at (axis cs:0, 0.5) {ATE};
      \node at (axis cs:1, 4) {BS};
      \node at (axis cs:1.5, 0) {CM};
      \node at (axis cs:6, 0) {CM};
      \node at (axis cs:3.5, 4) {P-CAL};
      \node at (axis cs:5, 2) {P-RES};
    \end{axis}
  \end{tikzpicture}
\end{subfigure}
\vspace{0.5cm}
\begin{subfigure}{0.45\textwidth}
    \subcaption{AUTOC}\label{subfigure:sim-heatmap-autoc}
    \begin{tikzpicture}[trim axis left]
    \begin{axis}[HeatmapNoiseFixed] 
      \addplot[
        matrix plot*, 
        mesh/rows=5,
        mesh/cols=9,
        point meta=explicit
      ] table[x=x, y=y, meta=cat_autoc] {\heatmap};
    \node at (axis cs:1.5, 4) {BS};
    \node at (axis cs:3.5, 0) {CM};
    \node at (axis cs:5, 2) {P-RES};
    \end{axis}
  \end{tikzpicture}
\end{subfigure}
\hfill
\begin{subfigure}{0.45\textwidth}
    \subcaption{Qini}\label{subfigure:sim-heatmap-qini}
    \begin{tikzpicture}[trim axis left]
    \begin{axis}[HeatmapNoiseFixed, ylabel={}] 
      \addplot[
        matrix plot*, 
        mesh/rows=5,
        mesh/cols=9,
        point meta=explicit
      ] table[x=x, y=y, meta=cat_qini] {\heatmap};
    \node at (axis cs:1.5, 4) {BS};
    \node at (axis cs:2, 0) {CM};
    \node at (axis cs:6.5, 0) {CM};
    \node at (axis cs:5, 2) {P-RES};
    \end{axis}
  \end{tikzpicture}
\end{subfigure}
\caption{    
\textbf{Best-performing methods across simulation scenarios.} 
Each panel shows the method with the best average performance for a given combination of base-score-to-CATE correlation ($\rho$) and training data size ($n$). Performance is evaluated for: \textbf{(a)} effect magnitude accuracy (MSE), \textbf{(b)} effect classification accuracy (EPG), and \textbf{(c–d)} effect ordering accuracy (AUTOC and Qini). The results illustrate the conceptual trade-off in Figure~\ref{fig:trade-off-analysis}: when $\rho$ is low, methods that make weaker assumptions about base scores (CM, P--COV) perform best; when $\rho$ is high and data is limited, simple approaches (BS, P--CAL) work best; and at intermediate correlations, residual post-processing (P--RES) often is most effective, especially for effect ordering.
}
\label{fig:sim-heatmap}
\end{figure}

\textbf{Weak alignment ($\rho\leq0.1$).} In this regime, the base scores are not helpful for modeling effect heterogeneity. Among the methods considered, only the ATE and the standard Causal Model (CM) do not rely on the base score. Consequently, CM generally dominates across all decision-making requirements. However, with very limited data it lacks the statistical power to outperform the simple ATE. For MSE this occurs when $n\leq 500$, and for EPG when $n\leq 250$.

\textbf{Data-scarce scenarios ($n\leq 500$).}
In settings with very limited experimental data, simpler methods are superior. For effect estimation, calibration post-processing (P--CAL) performs well by rescaling the score, though it requires strong alignment to surpass the ATE. For effect classification and ordering, base scores (BS) often perform best when alignment is medium to high ($\rho\geq 0.5$). This is because complex models (CM, P--RES, P--COV) are prone to overfitting and struggle to extract a reliable signal from limited data.

\textbf{Intermediate-to-high alignment with moderate data ($\rho\geq 0.3$, $n\geq 1000$).}
When the base score is informative but not perfect, with sufficient data the advantage shifts to more flexible post-processing methods. For ordering and classification, P--RES generally performs best, as it combines partial guidance from the score with corrections inferred from the experiment. For magnitude accuracy (MSE), results are mixed and highly sensitive to data availability. The pattern is consistent with Figure~\ref{fig:trade-off-analysis}: P--COV requires the largest samples to outperform alternatives, P--CAL performs best with smaller samples, and P--RES lies in between.

Overall, these results show that selecting the best approach is fundamentally an empirical task, shaped by the alignment of base scores with causal effects and the amount of experimental data available. Still, there is a clear pattern: when base scores are at least moderately informative, CPP tends to outperform conventional causal modeling---especially for effect ordering and classification.

\subsection{Cross-Validation for Choosing a Method in Practice}

The previous section showed which methods perform best under different alignment and data regimes. In practice, however, practitioners cannot directly measure alignment or individual effects, so they cannot simply look up the right approach.

Cross-validation (CV) offers a principled strategy to choose a method in practice~\citep{schaffer1993selecting}. For a collection of candidate algorithms $\mathcal{A}$, we can use CV to estimate each algorithm's expected performance and then choose the algorithm estimated to be best. The selected algorithm is 
\[
\hat{a} \;=\; \arg\max_{a \in \mathcal{A}}~ \widehat{\nu}(a),
\] 
where $\widehat{\nu}(a)$ is the CV estimate of performance (policy gain, reduction in MSE, or ordering metric, depending on the decision requirement). If CV is effective, $\hat{a}$ should approximate the oracle choice (see Appendix~\ref{App:sim-oracle} for a comparison against an Oracle benchmark, which confirms that the gap narrows quickly as data increase).

To assess the reliability of using CV to select CPP methods in practice, we evaluate two candidate sets. The \textbf{baseline group} reflects standard practice, including only the base-score (BS) and causal-model (CM) approaches. The \textbf{extended group} augments this set by adding the proposed CPP methods (P--CAL, P--RES, P--COV).

Recall that we run 100 simulations for each $(\rho,n)$ pair. Within each simulation, we use CV to select the best algorithm and its hyperparameters in both the baseline and extended groups (see Appendix~\ref{App:sim-cv} for details). We then compare their out-of-sample performance on a separate test set. To facilitate interpretation, we report the normalized improvement
\begin{equation}
\Delta(\rho,n) \;=\; 
\frac{\nu_{\text{ext}}(\rho,n) - \nu_{\text{base}}(\rho,n)}{\nu_{\text{ATE}}},
\label{eq:cv-improvement}
\end{equation}
where $\nu_{\text{ext}}(\rho,n)$ and $\nu_{\text{base}}(\rho,n)$ denote the out-of-sample performance of the algorithms selected by CV in the extended and baseline groups, respectively, and $\nu_{\text{ATE}}$ is the task-specific ATE benchmark (constant predictions for effect estimation, an ATE-based policy for classification, and a flat ranking for ordering). This normalization provides a common scale that is stable across $(\rho,n)$.

Figure~\ref{fig:sim-cv} summarizes the results. Two main patterns emerge.

\pgfplotstableread{simulation_figures/sim_cv_mse.dat}{\simcvmse}
\pgfplotstableread{simulation_figures/sim_cv_epg.dat}{\simcvepg}
\pgfplotstableread{simulation_figures/sim_cv_autoc.dat}{\simcvautoc}
\pgfplotstableread{simulation_figures/sim_cv_qini.dat}{\simcvqini}

\begin{figure}
  \centering
  \begin{subfigure}{0.48\textwidth}
    \subcaption{MSE}\label{subfig:sim-cv-mse}
    \begin{tikzpicture}[trim axis left]
      \begin{axis}[SimCVAxisStyle, xlabel={}]
        \addplot+[myCurveStyle] table[x=x, y=rho_0] {\simcvmse}; 
        \addlegendentry{$\rho=0.1$}
        \addplot+[myCurveStyle] table[x=x, y=rho_1] {\simcvmse}; 
        \addlegendentry{$\rho=0.3$}
        \addplot+[myCurveStyle] table[x=x, y=rho_2] {\simcvmse}; 
        \addlegendentry{$\rho=0.5$}
        \addplot+[myCurveStyle] table[x=x, y=rho_3] {\simcvmse}; 
        \addlegendentry{$\rho=0.7$}
        \addplot+[myCurveStyle] table[x=x, y=rho_4] {\simcvmse};
        \addlegendentry{$\rho=0.9$}
        \addplot[thick, black, dotted, mark=none] coordinates {(0, 0) (12, 0)};
        \node[below, font=\footnotesize] at (axis cs:11.5, 0) {ATE}; 
      \end{axis}
    \end{tikzpicture}
  \end{subfigure}
  \hfill
  \begin{subfigure}{0.48\textwidth}
    \subcaption{EPG}\label{subfig:sim-cv-epg}
    \begin{tikzpicture}[trim axis left]
      \begin{axis}[SimCVAxisStyle, xlabel={}, ylabel={}]
        \addplot+[myCurveStyle] table[x=x, y=rho_0] {\simcvepg}; 
        \addplot+[myCurveStyle] table[x=x, y=rho_1] {\simcvepg}; 
        \addplot+[myCurveStyle] table[x=x, y=rho_2] {\simcvepg};
        \addplot+[myCurveStyle] table[x=x, y=rho_3] {\simcvepg}; 
        \addplot+[myCurveStyle] table[x=x, y=rho_4] {\simcvepg};
        \addplot[thick, black, dotted, mark=none] coordinates {(0, 0) (12, 0)};
        \node[below, font=\footnotesize] at (axis cs:11.5, 0) {ATE};  
      \end{axis}
    \end{tikzpicture}
  \end{subfigure}
  \vspace{0.5cm} 

  \begin{subfigure}{0.48\textwidth}
    \subcaption{AUTOC}\label{subfig:sim-cv-autoc}
    \begin{tikzpicture}[trim axis left]
      \begin{axis}[SimCVAxisStyle]
        \addplot+[myCurveStyle] table[x=x, y=rho_0] {\simcvautoc}; 
        \addplot+[myCurveStyle] table[x=x, y=rho_1] {\simcvautoc}; 
        \addplot+[myCurveStyle] table[x=x, y=rho_2] {\simcvautoc}; 
        \addplot+[myCurveStyle] table[x=x, y=rho_3] {\simcvautoc}; 
        \addplot+[myCurveStyle] table[x=x, y=rho_4] {\simcvautoc};
        \addplot[thick, black, dotted, mark=none] coordinates {(0, 0) (12, 0)};
        \node[below, font=\footnotesize] at (axis cs:11.5, 0) {ATE}; 
      \end{axis}
    \end{tikzpicture}
  \end{subfigure}
  \hfill
  \begin{subfigure}{0.48\textwidth}
    \subcaption{Qini}\label{subfig:sim-cv-qini}
    \begin{tikzpicture}[trim axis left]
      \begin{axis}[SimCVAxisStyle, ylabel={}]
        \addplot+[myCurveStyle] table[x=x, y=rho_0] {\simcvqini}; 
        \addplot+[myCurveStyle] table[x=x, y=rho_1] {\simcvqini}; 
        \addplot+[myCurveStyle] table[x=x, y=rho_2] {\simcvqini}; 
        \addplot+[myCurveStyle] table[x=x, y=rho_3] {\simcvqini}; 
        \addplot+[myCurveStyle] table[x=x, y=rho_4] {\simcvqini}; 
        \addplot[thick, black, dotted] coordinates {(0, 0) (12, 0)};
        \node[below, font=\footnotesize] at (axis cs:11.5, 0) {ATE}; 
      \end{axis}
    \end{tikzpicture}
  \end{subfigure}
\caption{\textbf{Cross-validation analysis of the extended group's improvement over the baseline.} 
Each panel plots the normalized improvement $\Delta(\rho,n)$ (Eq.~\ref{eq:cv-improvement}), comparing the best algorithm selected by CV in the extended and baseline groups. Each line corresponds to a different alignment value ($\rho$). The results highlight two distinct patterns depending on the decision-making requirement: a monotonic decrease in benefit for magnitude estimation (a), and a non-monotonic ``rise-and-fall'' pattern for ordering and classification metrics (b--d).}

  \label{fig:sim-cv}
\end{figure}

For magnitude estimation (MSE), the extended group shows a clear advantage when alignment is moderate to high ($\rho \geq 0.5$), particularly with small to moderate data sizes. Because base scores do not provide meaningful estimates of effect magnitudes, the baseline group effectively reduces to a single candidate (CM), whereas the extended group includes multiple candidates (CM plus CPP methods). The figure highlights the greater data efficiency gained by allowing post-processing into the candidate set. For example, calibration (P--CAL) can rescale a strong but imperfect proxy with limited data, and CV reliably detects this advantage. As the sample size grows, however, CM gains sufficient statistical power to learn the CATE function directly from the experimental data, reducing and eventually eliminating the incremental benefit of considering multiple CPP candidates.

The dynamics differ for effect classification and ordering, where CPP's relative advantage follows a ``rise-and-fall'' pattern. With very limited data, CPP offers no benefit. Either the base score is uninformative (low $\rho$) or already strong (high $\rho$), in which case noisy post-processing is detrimental. The advantage peaks at moderate data sizes when alignment is medium to high. This is the ``sweet spot,'' where there is enough data to refine an imperfect yet informative base score, but still not enough for CM to fully recover the causal signal. When alignment is very high ($\rho=0.9$), the peak shifts to larger sample sizes because BS is a stronger contender, so more data is needed for CPP to add value beyond it. As $n$ grows further, CM continues to improve, and CPP’s advantage fades.

To unpack this rise-and-fall pattern, it helps to compare the extended set directly against each of the two baseline candidates. Figure~\ref{fig:sim-cv-qini-supp} shows that the preferred baseline method shifts with sample size. With very little data, BS is preferable because the experimental evidence is too scarce for CM to recover the causal signal. With abundant data, CM becomes a better option, as it has enough statistical power to estimate the CATE function reliably. The implication is that, as $n$ grows, the extended group consistently outperforms BS, but its advantage over CM narrows. CPP therefore adds the most value in the intermediate data regime.

\pgfplotstableread{simulation_figures/sim_cv_qini_bs.dat}{\simcvqinibs}
\pgfplotstableread{simulation_figures/sim_cv_qini_cm.dat}{\simcvqinicm}
\begin{figure}
  \centering
  \begin{subfigure}{0.48\textwidth}
    \subcaption{Qini: Extended vs. BS}\label{subfig:sim-cv-qini-bs}
    \begin{tikzpicture}[trim axis left]
      \begin{axis}[SimCVAxisStyle]
        \addplot+[myCurveStyle] table[x=x, y=rho_0] {\simcvqinibs}; 
        \addplot+[myCurveStyle] table[x=x, y=rho_1] {\simcvqinibs}; 
        \addplot+[myCurveStyle] table[x=x, y=rho_2] {\simcvqinibs}; 
        \addplot+[myCurveStyle] table[x=x, y=rho_3] {\simcvqinibs}; 
        \addplot+[myCurveStyle] table[x=x, y=rho_4] {\simcvqinibs};
        \addplot[thick, black, dotted, mark=none] coordinates {(0, 0) (12, 0)};
        \node[below, font=\footnotesize] at (axis cs:18, 0) {ATE}; 
      \end{axis}
    \end{tikzpicture}
  \end{subfigure}
  \hfill
  \begin{subfigure}{0.48\textwidth}
    \subcaption{Qini: Extended vs. CM}\label{subfig:sim-cv-qini-cm}
    \begin{tikzpicture}[trim axis left]
      \begin{axis}[SimCVAxisStyle, ylabel={}]
        \addplot+[myCurveStyle] table[x=x, y=rho_0] {\simcvqinicm}; 
        \addlegendentry{$\rho=0.1$}
        \addplot+[myCurveStyle] table[x=x, y=rho_1] {\simcvqinicm}; 
        \addlegendentry{$\rho=0.3$}
        \addplot+[myCurveStyle] table[x=x, y=rho_2] {\simcvqinicm};
        \addlegendentry{$\rho=0.5$}
        \addplot+[myCurveStyle] table[x=x, y=rho_3] {\simcvqinicm}; 
        \addlegendentry{$\rho=0.7$}
        \addplot+[myCurveStyle] table[x=x, y=rho_4] {\simcvqinicm};
        \addlegendentry{$\rho=0.9$}
        \addplot[thick, black, dotted, mark=none] coordinates {(0, 0) (12, 0)};
        \node[below, font=\footnotesize] at (axis cs:11.5, 0) {ATE};  
      \end{axis}
    \end{tikzpicture}
  \end{subfigure}
  \caption{\textbf{Cross-validation analysis for the Qini metric.} Each panel plots the normalized improvement $\Delta(\rho,n)$ (Eq.~\ref{eq:cv-improvement}) of the extended group relative to each baseline approach separately. The comparison reveals contrasting trends: the extended group’s advantage over BS tends to grow with data size (a), while its advantage over CM narrows (b).}
 
  \label{fig:sim-cv-qini-supp}
\end{figure}

Finally, note that the extended group also contains BS and CM. In principle, CV should select them when they are optimal. Why, then, can the baseline group sometimes do better? Because adding more candidates is not always helpful: when the CPP methods are not superior, CV may still select them due to sampling noise. In other words, CV is not infallible, and we see it mislead in some cases when alignment is very low or the experimental dataset is too small. However, the upside from including CPP in the candidate set is typically much larger than the downside, making the extended set the more reliable choice overall.

\subsection{Practical Implications for Experimentation}

A final consideration is the practical setting in which organizations actually run experiments. From a causal machine learning standpoint, it may seem that the real challenge is what to do once an intervention works---how to personalize it. But in practice, the bigger challenge comes before that: most experiments fail. The typical outcome of an online controlled experiment tends to be no measurable improvement or even a negative effect~\citep{kohavi2020trustworthy}.

This reality means that the path to finding genuinely valuable interventions is paved with many failed trials. Organizations therefore must run a steady stream of experiments, constantly testing new ideas. That necessity creates pressure on budgets and also introduces risks of confounding when many overlapping tests are conducted at once~\citep{abbasi2025critical}. As a result, the bottleneck is often not so much how to make the best possible personalization, but whether firms can efficiently identify interventions worth personalizing in the first place.

Suppose an intervention shows some evidence of effect. Even then, it does not follow that causal machine learning should be deployed to tailor actions across individuals. Several factors may block that step. Sometimes effects are nearly uniform, so a simple ATE policy is enough. Other times, heterogeneity exists but available features are not predictive of it, meaning no model can separate who benefits more or less. In other cases, there may be signal in principle but the experimental data are too limited to demonstrate it credibly. And even when heterogeneity is detectable, it may not be actionable unless it changes the optimal action for different subgroups~\citep{shchetkina2024heterogeneity}. In addition, the operational costs of building a system for individualized actions may outweigh the incremental benefit. In short, finding a promising intervention is necessary, but not sufficient, for personalization to be worthwhile.\footnote{Of course, an intervention need not have a positive ATE to be useful. If it benefits some individuals and harms others, personalization of intervention targeting can convert this variation into net value; see \citet{fernandez2023comparison} for a large-scale concrete case.}

This brings us back to the organizational constraints. Firms often need to try a lot of different things before they find interventions for which causal machine learning is worthwhile. And as firms scale up experimentation, interactions across concurrent experiments become increasingly likely to confound results, making biased inferences more likely~\citep{abbasi2025critical}. Thus, the need to control for confounding from concurrent experiments effectively constrains experimental budgets. Paradoxically, the problem is most acute for organizations that run many experiments. Because they test so many ideas, they cannot afford to devote massive samples to each one. Instead, they need to reach credible conclusions with as little data as possible. In other words, the strategic bottleneck is not the absence of personalization techniques, but the ability to evaluate interventions reliably and within budget.

With this backdrop, we use the simulation to illustrate the potential value of CPP, particularly in the common situation where many interventions are tested, most of which will not be relevant enough to justify personalization. To make this concrete, consider the case plotted in Figure~\ref{fig:sim-group}. Here we focus on an intervention that is worthy of personalization (given enough training data) with base-score alignment of 0.7. We compare three options: the extended group (BS + CM + CPP), the baseline group (BS + CM), and CM alone. In all cases performance improves with sample size, and with very large experiments the differences disappear---every approach eventually converges.

But in practice, organizations rarely run very large experiments when evaluating new interventions. Typically, they must try many different ideas under tight experimental budgets, which means they cannot afford to collect massive samples for each one. A common practice in experimentation is therefore to define a practical threshold: a minimum performance improvement that an idea must deliver before further investment is justified~\citep{kohavi2020trustworthy}. Suppose, for example, that for causal classification decisions the practical threshold is set such that personalization must achieve at least a 60\% improvement in expected policy gain (EPG) relative to an ATE-based policy.

The chart shows how CPP alters the picture. The extended group crosses this 60\% threshold at a data size of 5 (corresponding to 8,000 experimental observations), while the baseline group requires nearly 16,000 observations to reach the same level. In other words, CPP reduces the sample size needed by half. This means that, with the same experimental budget, an organization could test roughly twice as many interventions, substantially increasing its chances of discovering interventions that make personalization worthwhile.

The takeaway is that CPP is uniquely positioned to aid in the discovery of interventions worth personalizing. With very small data, personalized decisions are rarely credible, and with very large data, causal models alone eventually suffice. But to decide whether personalization is justified, organizations must operate in the intermediate regime---large enough to reveal the personalization potential if it exists, but not so large that scarce experimental budgets are exhausted on an intervention that may not be worth it. This moderate-data regime is exactly where CPP tends to shine, provided the base scores are at least moderately aligned with causal effects.

A final point is that considering both BS and CM without post-processing offers little benefit in this discovery problem. In our example, BS + CM performs no better than CM alone in this regard. This is because BS only provides an edge at very small data sizes, where personalization is unlikely to be practical in any case (unless BS is itself a very strong baseline, as we will see in the empirical study). Thus, the practical advantage comes not from simply pooling BS and CM, but from systematically integrating base scores through causal post-processing.

\pgfplotstableread{simulation_figures/sim_group_mse.dat}{\simgroupmse}
\pgfplotstableread{simulation_figures/sim_group_epg.dat}{\simgroupepg}
\pgfplotstableread{simulation_figures/sim_group_autoc.dat}{\simgroupautoc}
\pgfplotstableread{simulation_figures/sim_group_qini.dat}{\simgroupqini}

\begin{figure}
  \centering
  \begin{subfigure}{0.48\textwidth}
    \subcaption{MSE}\label{subfig:sim-group-mse}
    \begin{tikzpicture}[trim axis left]
      \begin{axis}[mySimAxisStyle, xlabel={}]
        \addplot+[myCurveStyle] table[x=x, y=cm] {\simgroupmse}; 
        \addplot+[myCurveStyle] table[x=x, y=base] {\simgroupmse}; 
        \addplot+[myCurveStyle] table[x=x, y=all] {\simgroupmse}; 
        \addplot[thick, black, dotted, mark=none] coordinates {(0, 0) (12, 0)};
        \node[below, font=\footnotesize] at (axis cs:11.5, 0) {ATE}; 
      \end{axis}
    \end{tikzpicture}
  \end{subfigure}
  \hfill
  \begin{subfigure}{0.48\textwidth}
    \subcaption{EPG}\label{subfig:sim-group-epg}
    \begin{tikzpicture}[trim axis left]
      \begin{axis}[mySimAxisStyle, xlabel={}, ylabel={}]
        \addplot+[myCurveStyle] table[x=x, y=cm] {\simgroupepg}; 
        \addplot+[myCurveStyle] table[x=x, y=base] {\simgroupepg}; 
        \addplot+[myCurveStyle] table[x=x, y=all] {\simgroupepg};
        \addplot[thick, black, dotted, mark=none] coordinates {(0, 0) (12, 0)};
        \node[below, font=\footnotesize] at (axis cs:11.5, 0) {ATE};  
      \end{axis}
    \end{tikzpicture}
  \end{subfigure}
  \vspace{0.5cm} 
  
  \begin{subfigure}{0.48\textwidth}
    \subcaption{AUTOC}\label{subfig:sim-group-autoc}
    \begin{tikzpicture}[trim axis left]
      \begin{axis}[mySimAxisStyle]
        \addplot+[myCurveStyle] table[x=x, y=cm] {\simgroupautoc}; 
        \addplot+[myCurveStyle] table[x=x, y=base] {\simgroupautoc}; 
        \addplot+[myCurveStyle] table[x=x, y=all] {\simgroupautoc}; 
        \addplot[thick, black, dotted, mark=none] coordinates {(0, 0) (12, 0)};
        \node[below, font=\footnotesize] at (axis cs:11.5, 0) {ATE}; 
      \end{axis}
    \end{tikzpicture}
  \end{subfigure}
  \hfill
  \begin{subfigure}{0.48\textwidth}
    \subcaption{Qini}\label{subfig:sim-group-qini}
    \begin{tikzpicture}[trim axis left]
      \begin{axis}[mySimAxisStyle, ylabel={}]
        \addplot+[myCurveStyle] table[x=x, y=cm] {\simgroupqini}; 
        \addlegendentry{\getLabel{CM}}
        \addplot+[myCurveStyle] table[x=x, y=base] {\simgroupqini}; 
        \addlegendentry{\getLabel{Baseline}}
        \addplot+[myCurveStyle] table[x=x, y=all] {\simgroupqini}; 
        \addlegendentry{\getLabel{Extended}}
        \addplot[thick, black, dotted] coordinates {(0, 0) (12, 0)};
        \node[below, font=\footnotesize] at (axis cs:11.5, 0) {ATE}; 
      \end{axis}
    \end{tikzpicture}
  \end{subfigure}
  \caption{\textbf{CPP improves statistical efficiency.} Under strong-but-imperfect alignment ($\rho=0.7$), the Extended group (CM + BS + CPP) reaches any given performance level with fewer observations than CM or the Baseline group (CM + BS).}
  \label{fig:sim-group}
\end{figure}

\section{Empirical Study}\label{sec:empirical}

Simulations clarified how alignment and data size shape method performance. We now ask whether these patterns carry over to a large, real-world setting with all its messiness (class imbalance, rare outcomes, and unknown ground truth). Using the Criteo advertising randomized experiment~\citep{diemert2018large}, we evaluate whether and when CPP improves targeting quality relative to conventional causal modeling and to using a proxy score alone.

Our empirical analysis serves three goals. First, it provides an external validity check: do the qualitative regimes from the simulations reappear on real data? Second, it examines operational constraints that simulations abstract away (e.g., the absence of an ``oracle'' base score). Third, it offers practical guidance on how to choose between methods.

The dataset comes from a randomized experiment where the treatment consists of ad exposure. Each entry corresponds to a user and includes 11 features, a treatment indicator, and two outcomes of interest: visit and conversion. The dataset contains 13,979,592 observations, with 85\% of users in the treated group. The visit rate is 4.70\%, the conversion rate is 0.29\%, the average treatment effect on visit is 1.03\%, and the average treatment effect on conversion is 0.12\%.

This study considers a hypothetical scenario in which a company aims to optimize its advertising budget to maximize purchases for a newly launched product. Because the product is new, only limited purchase data are available from an initial small-scale randomized experiment conducted to inform future ad targeting.

We study two types of proxy scores that are common in practice:
\begin{itemize}
    \item \textbf{Baseline visit predictions}: Companies often maintain predictive models of users’ likelihood of visiting a website absent advertising. While such models are not trained to predict purchases or behavior under treatment, they can still be useful for identifying individuals with potential interest in a product. Prior work has shown that website visits are strong proxies for conversions~\citep{dalessandro2015evaluating}. Thus, visit predictions may serve as an effective way to rank individuals by the potential impact of advertising. Importantly, such models are typically trained on large datasets and can be reused across multiple advertising campaigns.
    \item \textbf{Value metric}: Firms also rely on RFM (recency, frequency, and monetary value) metrics to segment customers~\citep{wei2010review}, for instance prioritizing those with recent or high-value purchases. Because the Criteo dataset anonymizes features and applies random projections, we cannot construct true RFM variables. Instead, we approximate an RFM-like metric using the feature most correlated with visits (f9). This serves as a proxy score that mimics how firms often use simple value-based measures in targeting.
\end{itemize}

These choices mirror the simulation’s ``alignment'' axis: as suggested by prior work, baseline-visit predictions are likely to align closely with conversion uplift, whereas a single value feature is less aligned but reflects the kinds of simple heuristics many organizations use. This also illustrates that CPP does not require base scores to come from a predictive model---it can be applied to any proxy and used to align it with experimental evidence.

\subsection{Empirical Setup}

We reserve 20\% of the data as a hold-out test set for evaluation. From the remaining sample, we draw 1,000,000 control-group observations to train the predictive model for baseline visits. In addition, we construct a small-scale randomized experiment data pool by sampling 250,000 observations evenly split between treatment and control (125,000 each). This experimental pool forms the basis for training data samples of varying sizes, ranging from 10,000 to 250,000 (in increments of 30,000). We repeat this process 100 times and report average results across replications.

The methods compared are the same as in Section~\ref{sec:exp_setup}, and performance is evaluated using the metrics defined in Section~\ref{sec:decision-making}: MSE for magnitude accuracy, EPG for classification accuracy, AUTOC and Qini for ordering accuracy.

Because individual-level CATEs are unobserved, we approximate MSE using the aggregation procedure of \citet{leng2024calibration}. Specifically, the test set is partitioned into 10 percentile bins based on the unprocessed base scores. For each bin $\mathcal{B}$, we compute the model-free treatment effect as the difference between average outcomes in the treatment and control groups, and compare it to the average predicted score. The overall MSE is the average squared error across bins:
\begin{align}
    \widehat{\text{MSE}} &= \frac{1}{|\mathcal{P}|}\sum_{\mathcal{B}\in \mathcal{P}}(\mu_{y}(1, \mathcal{B})-\mu_{y}(0, \mathcal{B})-\mu_{S}( \mathcal{B}))^2, \\
    \mu_{y}(t, \mathcal{B}) &= \frac{1}{N(t, \mathcal{B})}\sum_{i\in\mathcal{B}:t_i=t}y_i, \nonumber\\
    \mu_{S}(\mathcal{B}) &= \frac{1}{N(0,\mathcal{B})+N(1,\mathcal{B})}\sum_{i\in\mathcal{B}}S_i, \nonumber
\end{align}
where $\mathcal{P}$ is the set of bins. For each bin $\mathcal{B}$, $N(t,\mathcal{B})$ is the number of individuals with treatment $T=t$, $\mu_y(t,\mathcal{B})$ is the observed average outcome, and $\mu_{S}(\mathcal{B})$ is the average score. Each individual $i$ has a score $S_i\in\mathbb{R}$, outcome $y_i\in{0,1}$, and treatment assignment $t_i\in{0,1}$.

For classification, EPG is defined as the policy gain from targeting the top 10\% of individuals ranked by score. We evaluate this metric empirically by estimating the model-free average treatment effect among the top 10\% of scored individuals in the test set.

\subsection{Empirical Results}

Figures~\ref{fig:criteo-visit} and~\ref{fig:criteo-f9} present the empirical results for the two types of proxy scores. Together they illustrate both the strength and the limitations of proxy-based targeting, and how CPP can leverage such scores under realistic conditions.

\pgfplotstableread{criteo_figures/visit_mse.dat}{\visitmse}
\pgfplotstableread{criteo_figures/visit_epg.dat}{\visitepg}
\pgfplotstableread{criteo_figures/visit_autoc.dat}{\visitautoc}
\pgfplotstableread{criteo_figures/visit_qini.dat}{\visitqini}

\begin{figure}
  \centering
  \begin{subfigure}{0.48\textwidth}
    \subcaption{MSE}\label{subfig:criteo-visit-mse}
    \begin{tikzpicture}[trim axis left]
      \begin{axis}[myCriteoAxisStyle, xlabel={}]
        \addplot+[myCurveStyle] table[x=x, y=CF] {\visitmse}; 
        \addplot+[myCurveStyle] table[x=x, y=MC] {\visitmse}; 
        \addplot+[myCurveStyle] table[x=x, y=CRF] {\visitmse}; 
        \addplot+[myCurveStyle] table[x=x, y=CF-BS] {\visitmse}; 
        \addplot[thick, black, dotted, mark=none] coordinates {(0, 0) (25, 0)};
        \node[anchor=north, font=\footnotesize] at (axis cs:23.5, 0) {ATE}; 
      \end{axis}
    \end{tikzpicture}
  \end{subfigure}
  \hfill
  \begin{subfigure}{0.48\textwidth}
    \subcaption{EPG}\label{subfig:criteo-visit-epg}
    \begin{tikzpicture}[trim axis left]
      \begin{axis}[myCriteoAxisStyle, xlabel={}, ylabel={}]
        \addplot+[myCurveStyle] table[x=x, y=CF] {\visitepg}; 
        \addplot+[myCurveStyle] table[x=x, y=MC] {\visitepg}; 
        \addplot+[myCurveStyle] table[x=x, y=CRF] {\visitepg};
        \addplot+[myCurveStyle] table[x=x, y=CF-BS] {\visitepg}; 
        \addplot[thick, black, dotted, mark=none] coordinates {(0, 0) (25, 0)};
        \node[anchor=north, font=\footnotesize] at (axis cs:23.5, 0) {BS};  
      \end{axis}
    \end{tikzpicture}
  \end{subfigure}
  \vspace{0.5cm} 

  \begin{subfigure}{0.48\textwidth}
    \subcaption{AUTOC}\label{subfig:criteo-visit-autoc}
    \begin{tikzpicture}[trim axis left]
      \begin{axis}[myCriteoAxisStyle]
        \addplot+[myCurveStyle] table[x=x, y=CF] {\visitautoc}; 
        \addplot+[myCurveStyle] table[x=x, y=MC] {\visitautoc}; 
        \addplot+[myCurveStyle] table[x=x, y=CRF] {\visitautoc}; 
        \addplot+[myCurveStyle] table[x=x, y=CF-BS] {\visitautoc}; 
        \addplot[thick, black, dotted, mark=none] coordinates {(0, 0) (25, 0)};
        \node[anchor=north, font=\footnotesize] at (axis cs:23.5, 0) {BS}; 
      \end{axis}
    \end{tikzpicture}
  \end{subfigure}
  \hfill
  \begin{subfigure}{0.48\textwidth}
    \subcaption{Qini}\label{subfig:criteo-visit-qini}
    \begin{tikzpicture}[trim axis left]
      \begin{axis}[myCriteoAxisStyle, ylabel={}]
        \addplot+[myCurveStyle] table[x=x, y=CF] {\visitqini}; 
        \addlegendentry{\getLabel{CF}}
        \addplot+[myCurveStyle] table[x=x, y=MC] {\visitqini}; 
        \addlegendentry{\getLabel{MC}}
        \addplot+[myCurveStyle] table[x=x, y=CRF] {\visitqini}; 
        \addlegendentry{\getLabel{CRF}}
        \addplot+[myCurveStyle] table[x=x, y=CF-BS] {\visitqini}; 
        \addlegendentry{\getLabel{CF-BS}}
        \addplot[thick, black, dotted] coordinates {(0, 0) (25, 0)};
        \node[anchor=north, font=\footnotesize] at (axis cs:23.5, 0) {BS}; 
      \end{axis}
    \end{tikzpicture}
  \end{subfigure}
  \caption{
    \textbf{Empirical results using baseline visit predictions as base scores.} The raw base score (BS) is a remarkably strong benchmark for ordering and classification, outperforming most methods across a wide range of sample sizes. Residual post-processing (P--RES) performs best on AUTOC and EPG when the experimental data is sufficiently large. For magnitude accuracy (MSE), calibration post-processing (P--CAL) and the standard causal model (CM) perform best.}
  \label{fig:criteo-visit}
\end{figure}

\pgfplotstableread{criteo_figures/f9_mse.dat}{\rfmmse}
\pgfplotstableread{criteo_figures/f9_epg.dat}{\rfmepg}
\pgfplotstableread{criteo_figures/f9_autoc.dat}{\rfmautoc}
\pgfplotstableread{criteo_figures/f9_qini.dat}{\rfmqini}

\begin{figure}
  \centering
  \begin{subfigure}{0.48\textwidth}
    \subcaption{MSE}\label{subfig:criteo-f9-mse}
    \begin{tikzpicture}[trim axis left]
      \begin{axis}[myCriteoAxisStyle, xlabel={}]
        \addplot+[myCurveStyle] table[x=x, y=CF] {\rfmmse}; 
        \addplot+[myCurveStyle] table[x=x, y=MC] {\rfmmse}; 
        \addplot+[myCurveStyle] table[x=x, y=CRF] {\rfmmse}; 
        \addplot+[myCurveStyle] table[x=x, y=CF-BS] {\rfmmse}; 
        \addplot[thick, black, dotted, mark=none] coordinates {(0, 0) (25, 0)};
        \node[anchor=north, font=\footnotesize] at (axis cs:23.5, 0) {ATE}; 
      \end{axis}
    \end{tikzpicture}
  \end{subfigure}
  \hfill
  \begin{subfigure}{0.48\textwidth}
    \subcaption{EPG}\label{subfig:criteo-f9-epg}
    \begin{tikzpicture}[trim axis left]
      \begin{axis}[myCriteoAxisStyle, xlabel={}, ylabel={}]
        \addplot+[myCurveStyle] table[x=x, y=CF] {\rfmepg}; 
        \addplot+[myCurveStyle] table[x=x, y=MC] {\rfmepg}; 
        \addplot+[myCurveStyle] table[x=x, y=CRF] {\rfmepg};
        \addplot+[myCurveStyle] table[x=x, y=CF-BS] {\rfmepg}; 
        \addplot[thick, black, dotted, mark=none] coordinates {(0, 0) (25, 0)};
        \node[anchor=north, font=\footnotesize] at (axis cs:23.5, 0) {BS};  
      \end{axis}
    \end{tikzpicture}
  \end{subfigure}
  \vspace{0.5cm} 

  \begin{subfigure}{0.48\textwidth}
    \subcaption{AUTOC}\label{subfig:criteo-f9-autoc}
    \begin{tikzpicture}[trim axis left]
      \begin{axis}[myCriteoAxisStyle]
        \addplot+[myCurveStyle] table[x=x, y=CF] {\rfmautoc}; 
        \addplot+[myCurveStyle] table[x=x, y=MC] {\rfmautoc}; 
        \addplot+[myCurveStyle] table[x=x, y=CRF] {\rfmautoc}; 
        \addplot+[myCurveStyle] table[x=x, y=CF-BS] {\rfmautoc}; 
        \addplot[thick, black, dotted, mark=none] coordinates {(0, 0) (25, 0)};
        \node[anchor=north, font=\footnotesize] at (axis cs:23.5, 0) {BS}; 
      \end{axis}
    \end{tikzpicture}
  \end{subfigure}
  \hfill
  \begin{subfigure}{0.48\textwidth}
    \subcaption{Qini}\label{subfig:criteo-f9-qini}
    \begin{tikzpicture}[trim axis left]
      \begin{axis}[myCriteoAxisStyle, ylabel={}]
        \addplot+[myCurveStyle] table[x=x, y=CF] {\rfmqini}; 
        \addlegendentry{\getLabel{CF}}
        \addplot+[myCurveStyle] table[x=x, y=MC] {\rfmqini}; 
        \addlegendentry{\getLabel{MC}}
        \addplot+[myCurveStyle] table[x=x, y=CRF] {\rfmqini}; 
        \addlegendentry{\getLabel{CRF}}
        \addplot+[myCurveStyle] table[x=x, y=CF-BS] {\rfmqini}; 
        \addlegendentry{\getLabel{CF-BS}}
        \addplot[thick, black, dotted] coordinates {(0, 0) (25, 0)};
        \node[anchor=north, font=\footnotesize] at (axis cs:23.5, 0) {BS}; 
      \end{axis}
    \end{tikzpicture}
  \end{subfigure}
  \caption{
    \textbf{Empirical results using an RFM-like value metric (feature f9) as base scores.} Compared to baseline visit predictions, this single feature is less well aligned with conversion uplift. As a result, no post-processing method outperforms the standard causal model (CM) for magnitude accuracy (MSE). For ordering and classification, most methods eventually surpass the raw base score (BS), but only with substantial experimental data. Among them, residual post-processing (P--RES) delivers the greatest gains.}
  \label{fig:criteo-f9}
\end{figure}

A striking finding in Figure~\ref{fig:criteo-visit} is how difficult it is to improve upon the baseline visit score for prioritization (effect classification and ordering). This score simply predicts the likelihood of visiting a website in the absence of an ad. It is not a causal estimate. It does not even estimate the outcome of interest (conversions). Nevertheless, it aligns so strongly with conversion uplift that it serves as a formidable benchmark. For the Qini metric (Figure~\ref{subfig:criteo-visit-qini}), the raw base score (BS) outperforms all other methods across all sample sizes. For AUTOC and EPG (Figures~\ref{subfig:criteo-visit-autoc}, \ref{subfig:criteo-visit-epg}), BS remains dominant until there is a large enough experimental sample ($n \approx 70{,}000$), at which point residual post-processing (P--RES) begins to surpass it. Other methods eventually catch up as well, but only with substantially more data, underscoring just how difficult it is to beat a strong proxy score.

This difficulty underscores the need to think carefully about how proxy scores can be harnessed for causal decisions. Although BS is not causal per se, it nevertheless encodes a signal that reflects treatment effect heterogeneity---a signal that causal modeling alone cannot easily recover from limited, noisy experimental data. CPP offers a principled way to leverage this signal. Among the methods considered, P--RES is the only one that consistently surpasses BS on ordering and classification (given enough data), doing so by refining the proxy with evidence from the experiment.

For effect magnitude estimation (Figure~\ref{subfig:criteo-visit-mse}), the story is different. Here, all methods eventually improve upon the trivial ATE baseline. Interestingly, calibration post-processing (P--CAL) delivers the strongest improvements when data are not too scarce. Unlike other methods, P--CAL does not require access to any features---only to the base score itself. This makes it especially attractive in practical scenarios where feature data may be unavailable or costly to use. The standard causal model (CM) has similar performance, but the fact that such a simple score-based adjustment can outperform feature-rich methods underlines the opportunity that post-processing provides.

By contrast, Figure~\ref{fig:criteo-f9} shows what happens when the value metric (feature f9) is used as the proxy. Here the alignment with conversion uplift is weaker, so the preferred methods shift. First, no post-processing technique is able to outperform CM on MSE; in this case, modeling directly from features provides the most reliable effect estimates. Second, for ordering and classification, most methods eventually surpass BS, but only with substantial experimental data. Among them, P--RES still stands out as the most effective, consistently delivering the largest gains.

Taken together, the empirical results show that the value of post-processing depends on proxy quality and the decision requirement. With a strong proxy, P--CAL can provide surprisingly good effect magnitude estimates (assuming there is enough experimental data). With a weaker proxy, CPP doesn't improve magnitude estimation, but P--RES remains helpful for improving prioritization.

\section{Discussion}

Our findings provide a clear map for practitioners on how to best leverage Causal Post-Processing (CPP) based on their specific context. Several key themes emerge from our simulations and empirical analysis.

First, strong proxy scores can set a very high bar. They can perform remarkably well for effect ordering and classification tasks, sometimes leaving little room for causal modeling to add value unless experimental data are abundant. In these cases, post-processing is especially effective: rather than discarding the proxy and trying to estimate effect heterogeneity from scratch, it leverages the proxy's strength while incorporating experimental evidence to correct it (if necessary).

Second, the best method depends on proxy quality, data availability, and the decision requirement. When proxies are strongly aligned with treatment effects, calibration post-processing can provide accurate magnitude estimates with surprisingly little data, often outperforming more complex approaches without needing covariates. With weaker proxies, post-processing is less useful for magnitude estimation but can still be highly valuable for prioritization. In this regime, residual post-processing offers the largest gains, though it requires at least moderate alignment and sufficient experimental data. More broadly, our simulations reveal a clear pattern: causal modeling excels with weak proxies and abundant data; calibration shines for effect magnitude estimation when proxies are strong and data scarce.  Residual post-processing tends to dominate the middle ground for all four of the decision-making settings.

At the same time, these results come with limitations. While simulations map out the regimes where different methods should excel, it remains an empirical question how often such regimes occur in practice. In particular, the alignment between proxy scores and treatment effects, and the size and quality of available experimental data, will determine the extent to which post-processing is valuable. The bad news is that very few real-world data sets are publicly available (to our knowledge) for research on individualized treatment assignment. The good news is that, nonetheless, in practice organizations have such data and methods can be evaluated directly: holdout data can be used to compare methods out of sample, and learning curve analysis can guide decisions about whether additional data collection is likely to change the preferred approach.

\section{Conclusion}
In this paper, we introduced causal post-processing (CPP) as a general framework for refining predictive scores with experimental data to improve intervention decisions. Our empirical results demonstrate that a well-chosen chosen CPP method can substantially improve decision quality over both conventional causal modeling and the direct use of proxy scores, particularly when a predictive score is informative of treatment effects but imperfectly aligned with them.

Our analysis highlights a key trade-off within the CPP framework. Different methods rest on distinct assumptions about how predictive scores relate to causal effects, which in turn shape their flexibility and data demands. The residual-based approach we proposed emerges as an effective compromise, offering strong performance in intermediate-data regimes by balancing adaptability with data efficiency. We show how this advantage can potentially help organizations increase their chances of discovering interventions that make personalization worthwhile.

The modular nature of the CPP framework offers substantial practical advantages. It builds upon existing predictive scores, which are often trained on large-scale observational data, enabling organizations to repurpose existing predictive assets for causal tasks. CPP also is well suited for cases where the same predictive model (scores) may be reused across multiple interventions. For example, many targeting applications---including advertising and churn management—involve many sequential campaigns with different incentives and creatives. Each such campaign begins with a cold-start problem to which CPP provides a potential data-efficient solution. 

This work opens several promising directions for future research. A natural next step is to extend residual post-processing beyond forest-based learners to other flexible estimators such as BART, which often outperform causal forests in empirical benchmarks~\citep{dorie2019automated}. Because BART can estimate individual-level effects with uncertainty, it may also enable finer-tuned calibration post-processing without the coarsening involved in quantile splitting (for a related example of BART-based post-processing, see \citealp{carn:dori:hill:2019}). Our framework assumes that the base score is given, making it applicable even when the underlying data or model internals are inaccessible. When those data are available, incorporating them directly into causal estimation rather than relying solely on derived scores is a promising extension. More broadly, developing CPP variants tailored to specific decision contexts presents a rich avenue for future research.

%
%
%

\section*{APPENDIX}
\appendix

\section{Proof that ranking scoring functions based on Equation~\eqref{eq:epg} is equivalent to ranking them based on Equation~\eqref{eq:ewm}}\label{App:ewm}
Minimizing EWM corresponds to
\begin{align*}
&\argmin_{S}\mathbb{E}[|\tau-\kappa|\cdot\mathbf{1}\{\tau>\kappa\neq S>\tilde\kappa\}]\\
=&\argmin_{S}\mathbb{E}[(\tau-\kappa)\cdot(\mathbf{1}\{\tau>\kappa\}-\mathbf{1}\{S>\tilde\kappa\})]\\
=&\argmin_{S}\mathbb{E}[(\tau-\kappa)\cdot\mathbf{1}\{\tau>\kappa\}]-\mathbb{E}[(\tau-\kappa)\cdot\mathbf{1}\{S>\tilde\kappa\}]
\end{align*}
Because the first term does not affect how scoring functions are ranked, it can be removed:
\begin{align*}
    =&\argmax_{S}\mathbb{E}[(\tau-\kappa)\cdot\mathbf{1}\{S>\tilde\kappa\}]
\end{align*}
Apply definition in Equation~\ref{eq:action}:
\begin{align*}
    =&\argmax_{S}\mathbb{E}[(\tau-\kappa)\cdot A]   \qed
\end{align*}

\section{Simulation Study}\label{app:extended}
This appendix supplements the simulation study by providing: 
\begin{enumerate}
    \item Results on the best-performing methods under a fixed data size with varying noise levels.
    \item Details of the cross-validation strategy.
    \item A comparison of the performance between Oracle and cross-validation–selected models.
\end{enumerate}

\subsection{Best-performing Methods with Varying Noise Levels}

Figure~\ref{fig:sim-heatmap-app} presents the best-performing methods at a fixed data size ($n=1000$), examining how proxy alignment ($\rho$) and noise ($\sigma_\epsilon^2$) interact. The results mirror the general patterns in Figure~\ref{fig:sim-heatmap}, where noise is fixed and data size varies, because both data size and noise levels influence how hard it is to model effect heterogeneity.

Across both analyses, the same trade-off holds true:
\begin{itemize}
    \item When the signal is weak (due to high noise or small data), simpler methods such as ATE, BS and P--CAL are favored, as they are less prone to overfitting.
    \item When the signal is strong (low noise or a large data), more flexible methods such as P--RES or P--COV perform best, as sufficient data is available to capture complex patterns.
    \item Alignment plays a key role in differentiating methods. ATE, CM, and P--COV rely the least on base scores, so they tend to work better when alignment is weak. 
\end{itemize}

Although both low noise and large datasets improve estimation, they affect model performance differently. Low noise improves signal strength, while larger datasets provide the statistical power for flexible models to learn complex patterns. This distinction leads to two results that differ from those shown in Figure~\ref{fig:sim-heatmap}:
\begin{itemize}
    \item When alignment is high ($\rho=0.9$), simpler methods like BS or P--CAL dominate regardless of signal strength. The fixed data size prevents more flexible methods like P--RES from gaining enough statistical power to outperform them, in contrast to what happened in the large-data scenarios of Figure~\ref{fig:sim-heatmap}.
    \item Even when alignment is very low ($\rho=0.1$), CM may still not be the best choice because the data is insufficient to fully learn the CATE function.
\end{itemize}

\pgfplotstableread{./simulation_figures/heatmap_datasize_fixed.dat}{\heatmapapp}

\begin{figure}
\centering
\begin{subfigure}{0.45\textwidth}
    \subcaption{MSE}
    \begin{tikzpicture}[trim axis left]
        \begin{axis}[HeatmapDataSizeFixed, xlabel={}] 
          \addplot[
            matrix plot*, 
            mesh/rows=5,
            mesh/cols=7,
            point meta=explicit
          ] table[x=x, y=y, meta=cat_mse] {\heatmapapp};
          \node at (axis cs:0.5, 1) {ATE};
          \node at (axis cs:3, 0) {CM};
          \node at (axis cs:2, 4) {P-CAL};
          \node at (axis cs:5, 3) {P-RES};
          \node at (axis cs:5, 1) {P-COV};
        \end{axis}
    \end{tikzpicture}
\end{subfigure}
\hfill
\begin{subfigure}{0.45\textwidth}
    \subcaption{EPG}
    \begin{tikzpicture}[trim axis left]
    \begin{axis}[HeatmapDataSizeFixed, xlabel={},ylabel={}] 
      \addplot[
        matrix plot*, 
        mesh/rows=5,
        mesh/cols=7,
        point meta=explicit
      ] table[x=x, y=y, meta=cat_epg] {\heatmapapp};
      \node at (axis cs:0, 0.5) {ATE};
      \node at (axis cs:1, 4) {BS};
      \node at (axis cs:1.5, 0) {CM};
      \node at (axis cs:4.5, 4) {P-CAL};
      \node at (axis cs:3.5, 2) {P-RES};
      \node at (axis cs:5.5, 0) {P-COV};
    \end{axis}
  \end{tikzpicture}
\end{subfigure}
\vspace{0.5cm}
\begin{subfigure}{0.45\textwidth}
    \subcaption{AUTOC}
    \begin{tikzpicture}[trim axis left]
        \begin{axis}[HeatmapDataSizeFixed] 
          \addplot[
            matrix plot*, 
            mesh/rows=5,
            mesh/cols=7,
            point meta=explicit
          ] table[x=x, y=y, meta=cat_autoc] {\heatmapapp};
          \node at (axis cs:3, 4) {BS};
          \node at (axis cs:2, 0) {CM};
          \node at (axis cs:3.5, 2) {P-RES};
          \node[font=\footnotesize] at (axis cs:6, 0) {P-COV};
        \end{axis}
    \end{tikzpicture}
\end{subfigure}
\hfill
\begin{subfigure}{0.45\textwidth}
    \subcaption{Qini}
    \begin{tikzpicture}[trim axis left]
    \begin{axis}[HeatmapDataSizeFixed, ylabel={}] 
      \addplot[
        matrix plot*, 
        mesh/rows=5,
        mesh/cols=7,
        point meta=explicit
      ] table[x=x, y=y, meta=cat_qini] {\heatmapapp};
      \node at (axis cs:3, 4) {BS};
      \node at (axis cs:2, 0) {CM};
      \node at (axis cs:3.5, 2) {P-RES};
      \node[font=\footnotesize] at (axis cs:6, 0) {P-COV};
    \end{axis}
  \end{tikzpicture}
\end{subfigure}
\caption{\textbf{Best-performing methods with varying noise levels.} This analysis confirms the same fundamental trade-off seen in Figure~\ref{fig:sim-heatmap}: simple methods (BS, P--CAL) excel in low signal-to-noise (SNR) settings, while flexible methods (P--RES, P--COV) are superior when the signal is strong. However, it also highlights a key distinction: even with low noise, flexible methods do not achieve the same dominance as they do with large datasets, particularly when alignment is high.}
\label{fig:sim-heatmap-app}
\end{figure}

\subsection{Cross-Validation Strategy}\label{App:sim-cv}

We adopt a hybrid cross-validation (CV) strategy that scales with training data size to balance accuracy and computational cost:

\begin{itemize}
\item \textbf{Small datasets} ($n \leq 8{,}000$): For each simulation, perform a full 10-repeated 5-fold CV to jointly select the best method and its optimal hyperparameters.
\item \textbf{Medium datasets} ($16{,}000 \leq n \leq 64{,}000$): For each simulation, use pre-tuned hyperparameters for each method and run a 5-fold CV to select only the best-performing method.
\item \textbf{Large datasets} ($n \geq 128{,}000$): Use pre-tuned parameter for each method and apply a single train–validation split to select the best-performing method.
\end{itemize}

Given that the true CATE is unknown, we use the following proxy criteria for model selection in different contexts:
\begin{itemize}
    \item Magnitude Estimation (MSE): The squared error between the transformed outcome $y_i^*$ and the CATE prediction $\hat\tau_i$.
    \[
        \frac{1}{n}\sum_{i=1}^n(y^*_i-\hat{\tau})^2,\quad\text{where }y_i^*=y_i\cdot\frac{t_i-p_t}{p_t(1-p_t)}
    \]
    and $p_t$ is the treated proportion.
    \item Effect classification (EPG): An unbiased estimator of the policy value $\mathbb{E}[Y^a]$.
    \[
        \frac{1}{n}\sum_{i=1}^n\frac{y_i}{p(t_i)}\cdot\mathbf{1}(t_i=a_i)
    \]
    where $p(t_i)$ is the known randomization probability for $T=t_i$ . 
    \item Effect ordering (AUTOC and Qini): The TOC is estimated using the difference in empirical outcome means for targeted subgroups, rather than relying on true CATE values.
\end{itemize}

When reporting the results, the improvement of the extended group over the baseline is normalized by the performance of a task-specific ATE benchmark, as shown in Equation~\ref{eq:cv-improvement}. The ATE baseline performance for each metric is given by:
\begin{itemize}
    \item \textbf{MSE:} The variance of the true CATE. $\text{MSE}_{\text{ATE}}=\sigma_{\Delta}^2$.
    \item \textbf{EPG:} The value of a policy that treats every individual. $\text{EPG}_{\text{ATE}}=\mu_{\Delta}$.
    \item \textbf{AUTOC and Qini}: The expected performance by ranking individuals at random.
    \[
        \int_0^1\alpha(q)\cdot\mu_{\Delta}~dq
    \]
    with $\alpha(q)=1$ for AUTOC and $\alpha(q)=q$ for Qini. It follows that, $\text{AUTOC}_{\text{ATE}}=\mu_{\Delta}$ and $\text{Qini}_{\text{ATE}}=0.5\mu_{\Delta}$.
\end{itemize}

\subsection{Comparison of Oracle and Cross-Validation Selection} \label{App:sim-oracle}

To assess the efficacy of our cross-validation (CV) procedure, we compare it against a theoretical Oracle selector. The Oracle serves as a benchmark of ideal performance, as it has access to the test set and the true CATE. 

For each experimental setup (i.e. each combination of alignment $\rho$, training size, and decision-making context), the Oracle identifies a single algorithm configuration: the one that achieves the optimal average test performance across all 100 simulations. In contrast, our practical CV procedure selects a model independently within each simulation, and its performance is the average of these 100 (potentially different) selected algorithms. We define the Oracle this way to avoid leakage---that is, to prevent the sampling distribution of the test data from influencing model selection, which would artificially inflate performance estimates. By choosing the algorithm that performs best on average across all simulations, the Oracle represents a stable upper benchmark of generalization rather than overfitting to any specific test set.

The resulting gap between the Oracle's performance and the CV-selected performance represents the potential performance lost to model selection uncertainty on a finite sample. Figure~\ref{fig:sim-cv-ora} plots this performance gap, normalized by the performance of an ATE model:
$$
\frac{\nu_{\text{ora}}-\nu_{\text{cv}}}{\nu_{\text{ATE}}}\cdot 100
$$
where $\nu_{\text{ora}}$, $\nu_{\text{cv}}$, and $\nu_{\text{ATE}}$ denote the performance values for the Oracle, CV-selected, and ATE models, respectively. As shown, the gap is substantial when the training data is scarce ($n\leq 1000$), with the highest alignment value settings ($\rho=0.9$) showing the largest discrepancy. This occurs because CV procedure struggles to identify the optimal model under high sampling noise, and the cost of missing that is higher when alignment is high. As the training data size increases, the CV estimates become more reliable, and the gap diminishes rapidly, approaching zero. 

\pgfplotstableread{simulation_figures/cv_vs_ora_mse.dat}{\cvoramse}
\pgfplotstableread{simulation_figures/cv_vs_ora_epg.dat}{\cvoraepg}
\pgfplotstableread{simulation_figures/cv_vs_ora_autoc.dat}{\cvoraautoc}
\pgfplotstableread{simulation_figures/cv_vs_ora_qini.dat}{\cvoraqini}

\begin{figure}
  \centering
  \begin{subfigure}{0.48\textwidth}
    \subcaption{MSE}
    \begin{tikzpicture}[trim axis left]
      \begin{axis}[SimCVAxisStyle, xlabel={}]
        \addplot+[myCurveStyle] table[x=x, y=rho_0] {\cvoramse}; 
        \addlegendentry{$\rho=0.1$}
        \addplot+[myCurveStyle] table[x=x, y=rho_1] {\cvoramse}; 
        \addlegendentry{$\rho=0.3$}
        \addplot+[myCurveStyle] table[x=x, y=rho_2] {\cvoramse}; 
        \addlegendentry{$\rho=0.5$}
        \addplot+[myCurveStyle] table[x=x, y=rho_3] {\cvoramse}; 
        \addlegendentry{$\rho=0.7$}
        \addplot+[myCurveStyle] table[x=x, y=rho_4] {\cvoramse};
        \addlegendentry{$\rho=0.9$}
        \addplot[thick, black, dotted, mark=none] coordinates {(0, 0) (12, 0)};
        \node[below, font=\footnotesize] at (axis cs:11.5, 0) {ATE}; 
      \end{axis}
    \end{tikzpicture}
  \end{subfigure}
  \hfill
  \begin{subfigure}{0.48\textwidth}
    \subcaption{EPG}
    \begin{tikzpicture}[trim axis left]
      \begin{axis}[SimCVAxisStyle, xlabel={}, ylabel={}]
        \addplot+[myCurveStyle] table[x=x, y=rho_0] {\cvoraepg}; 
        \addplot+[myCurveStyle] table[x=x, y=rho_1] {\cvoraepg}; 
        \addplot+[myCurveStyle] table[x=x, y=rho_2] {\cvoraepg};
        \addplot+[myCurveStyle] table[x=x, y=rho_3] {\cvoraepg}; 
        \addplot+[myCurveStyle] table[x=x, y=rho_4] {\cvoraepg};
        \addplot[thick, black, dotted, mark=none] coordinates {(0, 0) (12, 0)};
        \node[below, font=\footnotesize] at (axis cs:11.5, 0) {ATE};  
      \end{axis}
    \end{tikzpicture}
  \end{subfigure}
  \vspace{0.5cm} 

  \begin{subfigure}{0.48\textwidth}
    \subcaption{AUTOC}
    \begin{tikzpicture}[trim axis left]
      \begin{axis}[SimCVAxisStyle]
        \addplot+[myCurveStyle] table[x=x, y=rho_0] {\cvoraautoc}; 
        \addplot+[myCurveStyle] table[x=x, y=rho_1] {\cvoraautoc}; 
        \addplot+[myCurveStyle] table[x=x, y=rho_2] {\cvoraautoc}; 
        \addplot+[myCurveStyle] table[x=x, y=rho_3] {\cvoraautoc}; 
        \addplot+[myCurveStyle] table[x=x, y=rho_4] {\cvoraautoc};
        \addplot[thick, black, dotted, mark=none] coordinates {(0, 0) (12, 0)};
        \node[below, font=\footnotesize] at (axis cs:11.5, 0) {ATE}; 
      \end{axis}
    \end{tikzpicture}
  \end{subfigure}
  \hfill
  \begin{subfigure}{0.48\textwidth}
    \subcaption{Qini}
    \begin{tikzpicture}[trim axis left]
      \begin{axis}[SimCVAxisStyle, ylabel={}]
        \addplot+[myCurveStyle] table[x=x, y=rho_0] {\cvoraqini}; 
        \addplot+[myCurveStyle] table[x=x, y=rho_1] {\cvoraqini}; 
        \addplot+[myCurveStyle] table[x=x, y=rho_2] {\cvoraqini}; 
        \addplot+[myCurveStyle] table[x=x, y=rho_3] {\cvoraqini}; 
        \addplot+[myCurveStyle] table[x=x, y=rho_4] {\cvoraqini}; 
        \addplot[thick, black, dotted] coordinates {(0, 0) (12, 0)};
        \node[below, font=\footnotesize] at (axis cs:11.5, 0) {ATE}; 
      \end{axis}
    \end{tikzpicture}
  \end{subfigure}
\caption{\textbf{Performance Gap Between Oracle and CV-Selected Models.} 
Each panel displays the performance advantage of the Oracle over CV, normalized by the performance of a baseline ATE model. The gap is most pronounced in low-data regimes ($N\leq 1000$) but rapidly approaches zero as the training sample size increases, allowing CV to more reliably identify the optimal model.}
  \label{fig:sim-cv-ora}
\end{figure}

\section*{Acknowledgements}

This work was supported by the Research Grants Council [grant number 26500822].  Foster Provost thanks Ira Rennert and the NYU/Stern Fubon Center for support.


\bibliographystyle{informs2014} 
\bibliography{reference} 

\begin{thebibliography}{34}
\providecommand{\natexlab}[1]{#1}
\providecommand{\url}[1]{\texttt{#1}}
\providecommand{\urlprefix}{URL }

\bibitem[{Abbasi et~al.(2025)Abbasi, Somanchi, \protect\BIBand{} Kelley}]{abbasi2025critical}
Abbasi A, Somanchi S, Kelley K (2025) The critical challenge of using large-scale digital experiment platforms for scientific discovery. \emph{MIS Quarterly} 49(1):1--28.

\bibitem[{Ascarza et~al.(2018)Ascarza, Neslin, Netzer, Anderson, Fader, Gupta, Hardie, Lemmens, Libai, Neal et~al.}]{ascarza2018pursuit}
Ascarza E, Neslin SA, Netzer O, Anderson Z, Fader PS, Gupta S, Hardie BG, Lemmens A, Libai B, Neal D, et~al. (2018) In pursuit of enhanced customer retention management: Review, key issues, and future directions. \emph{Customer Needs and Solutions} 5:65--81.

\bibitem[{Athey \protect\BIBand{} Imbens(2016)}]{athey2016recursive}
Athey S, Imbens G (2016) Recursive partitioning for heterogeneous causal effects. \emph{Proceedings of the National Academy of Sciences} 113(27):7353--7360.

\bibitem[{Athey et~al.(2025)Athey, Keleher, \protect\BIBand{} Spiess}]{athey2025machine}
Athey S, Keleher N, Spiess J (2025) Machine learning who to nudge: causal vs predictive targeting in a field experiment on student financial aid renewal. \emph{Journal of Econometrics} 249:105945.

\bibitem[{{Carnegie} et~al.(2019){Carnegie}, {Dorie}, \protect\BIBand{} {Hill}}]{carn:dori:hill:2019}
{Carnegie} N, {Dorie} V, {Hill} J (2019) {Examining treatment effect heterogeneity using BART}. \emph{Observational Studies} 4.

\bibitem[{Dalessandro et~al.(2015)Dalessandro, Hook, Perlich, \protect\BIBand{} Provost}]{dalessandro2015evaluating}
Dalessandro B, Hook R, Perlich C, Provost F (2015) Evaluating and optimizing online advertising: Forget the click, but there are good proxies. \emph{Big data} 3(2):90--102.

\bibitem[{Devriendt et~al.(2020)Devriendt, Van~Belle, Guns, \protect\BIBand{} Verbeke}]{devriendt2020learning}
Devriendt F, Van~Belle J, Guns T, Verbeke W (2020) Learning to rank for uplift modeling. \emph{IEEE Transactions on Knowledge and Data Engineering} 34(10):4888--4904.

\bibitem[{Diemert et~al.(2018)Diemert, Betlei, Renaudin, \protect\BIBand{} Amini}]{diemert2018large}
Diemert E, Betlei A, Renaudin C, Amini MR (2018) A large scale benchmark for uplift modeling. \emph{KDD}.

\bibitem[{Dorie et~al.(2019)Dorie, Hill, Shalit, Scott, \protect\BIBand{} Cervone}]{dorie2019automated}
Dorie V, Hill J, Shalit U, Scott M, Cervone D (2019) Automated versus do-it-yourself methods for causal inference: Lessons learned from a data analysis competition. \emph{Statistical Science} 34(1):43--68.

\bibitem[{Fern{\'a}ndez-Lor{\'\i}a \protect\BIBand{} Provost(2022{\natexlab{a}})}]{fernandez2022causal}
Fern{\'a}ndez-Lor{\'\i}a C, Provost F (2022{\natexlab{a}}) Causal classification: Treatment effect estimation vs. outcome prediction. \emph{Journal of Machine Learning Research} 23(59):1--35.

\bibitem[{Fern{\'a}ndez-Lor{\'\i}a \protect\BIBand{} Provost(2022{\natexlab{b}})}]{fernandez2022causaldecision}
Fern{\'a}ndez-Lor{\'\i}a C, Provost F (2022{\natexlab{b}}) Causal decision making and causal effect estimation are not the same… and why it matters. \emph{INFORMS Journal on Data Science} 1(1):4--16.

\bibitem[{Fern{\'a}ndez-Lor{\'\i}a \protect\BIBand{} Provost(2025)}]{fernandez2025observational}
Fern{\'a}ndez-Lor{\'\i}a C, Provost F (2025) Observational vs. experimental data when making automated decisions using machine learning. \emph{INFORMS Journal on Data Science} .

\bibitem[{Fern{\'a}ndez-Lor{\'\i}a et~al.(2023)Fern{\'a}ndez-Lor{\'\i}a, Provost, Anderton, Carterette, \protect\BIBand{} Chandar}]{fernandez2023comparison}
Fern{\'a}ndez-Lor{\'\i}a C, Provost F, Anderton J, Carterette B, Chandar P (2023) A comparison of methods for treatment assignment with an application to playlist generation. \emph{Information Systems Research} 34(2):786--803.

\bibitem[{Gao et~al.(2024)Gao, Zheng, Wang, Feng, He, \protect\BIBand{} Li}]{gao2024causal}
Gao C, Zheng Y, Wang W, Feng F, He X, Li Y (2024) Causal inference in recommender systems: A survey and future directions. \emph{ACM Transactions on Information Systems} 42(4):1--32.

\bibitem[{Gordon et~al.(2023)Gordon, Moakler, \protect\BIBand{} Zettelmeyer}]{gordon2023close}
Gordon BR, Moakler R, Zettelmeyer F (2023) Close enough? a large-scale exploration of non-experimental approaches to advertising measurement. \emph{Marketing Science} 42(4):768--793.

\bibitem[{Hansen(2008)}]{hansen2008prognostic}
Hansen BB (2008) The prognostic analogue of the propensity score. \emph{Biometrika} 95(2):481--488.

\bibitem[{Hill(2011)}]{Hill2011Bayesian}
Hill JL (2011) Bayesian nonparametric modeling for causal inference. \emph{Journal of Computational and Graphical Statistics} 20:217 -- 240, \urlprefix\url{https://api.semanticscholar.org/CorpusID:122155840}.

\bibitem[{Kent et~al.(2020)Kent, Paulus, Van~Klaveren, D'Agostino, Goodman, Hayward, Ioannidis, Patrick-Lake, Morton, Pencina et~al.}]{kent2020predictive}
Kent DM, Paulus JK, Van~Klaveren D, D'Agostino R, Goodman S, Hayward R, Ioannidis JP, Patrick-Lake B, Morton S, Pencina M, et~al. (2020) The predictive approaches to treatment effect heterogeneity (path) statement. \emph{Annals of internal medicine} 172(1):35--45.

\bibitem[{Kohavi et~al.(2020)Kohavi, Tang, \protect\BIBand{} Xu}]{kohavi2020trustworthy}
Kohavi R, Tang D, Xu Y (2020) \emph{Trustworthy online controlled experiments: A practical guide to a/b testing} (Cambridge University Press).

\bibitem[{Lee et~al.(2020)Lee, Gopal, \protect\BIBand{} Park}]{lee2020different}
Lee D, Gopal A, Park SH (2020) Different but equal? a field experiment on the impact of recommendation systems on mobile and personal computer channels in retail. \emph{Information Systems Research} 31(3):892--912.

\bibitem[{Leng \protect\BIBand{} Dimmery(2024)}]{leng2024calibration}
Leng Y, Dimmery D (2024) Calibration of heterogeneous treatment effects in randomized experiments. \emph{Information Systems Research} .

\bibitem[{Li et~al.(2022)Li, Grahl, \protect\BIBand{} Hinz}]{li2022recommender}
Li X, Grahl J, Hinz O (2022) How do recommender systems lead to consumer purchases? a causal mediation analysis of a field experiment. \emph{Information Systems Research} 33(2):620--637.

\bibitem[{McFowland~III et~al.(2021)McFowland~III, Gangarapu, Bapna, \protect\BIBand{} Sun}]{mcfowland2021prescriptive}
McFowland~III E, Gangarapu S, Bapna R, Sun T (2021) A prescriptive analytics framework for optimal policy deployment using heterogeneous treatment effects. \emph{MIS Quarterly} 45(4).

\bibitem[{Nie \protect\BIBand{} Wager(2021)}]{nie2021quasi}
Nie X, Wager S (2021) Quasi-oracle estimation of heterogeneous treatment effects. \emph{Biometrika} 108(2):299--319.

\bibitem[{Peysakhovich \protect\BIBand{} Lada(2016)}]{peysakhovich2016combining}
Peysakhovich A, Lada A (2016) Combining observational and experimental data to find heterogeneous treatment effects. \emph{arXiv preprint arXiv:1611.02385} .

\bibitem[{Rosenbaum \protect\BIBand{} Rubin(1983)}]{rosenbaum1983central}
Rosenbaum PR, Rubin DB (1983) The central role of the propensity score in observational studies for causal effects. \emph{Biometrika} 70(1):41--55.

\bibitem[{Rubin(1990)}]{rubin1990comment}
Rubin DB (1990) Comment: Neyman (1923) and causal inference in experiments and observational studies. \emph{Statistical Science} 5(4):472--480.

\bibitem[{Schaffer(1993)}]{schaffer1993selecting}
Schaffer C (1993) Selecting a classification method by cross-validation. \emph{Machine learning} 13(1):135--143.

\bibitem[{Shchetkina \protect\BIBand{} Berman(2024)}]{shchetkina2024heterogeneity}
Shchetkina A, Berman R (2024) When is heterogeneity actionable for targeting? \emph{Proceedings of the 25th ACM Conference on Economics and Computation}, 778--779.

\bibitem[{Stitelman et~al.(2011)Stitelman, Dalessandro, Perlich, \protect\BIBand{} Provost}]{stitelman2011estimating}
Stitelman O, Dalessandro B, Perlich C, Provost F (2011) Estimating the effect of online display advertising on browser conversion. \emph{Data Mining and Audience Intelligence for Advertising (ADKDD 2011)} 8.

\bibitem[{Wager \protect\BIBand{} Athey(2018)}]{wager2018estimation}
Wager S, Athey S (2018) Estimation and inference of heterogeneous treatment effects using random forests. \emph{Journal of the American Statistical Association} 113(523):1228--1242.

\bibitem[{Wei et~al.(2010)Wei, Lin, \protect\BIBand{} Wu}]{wei2010review}
Wei JT, Lin SY, Wu HH (2010) A review of the application of rfm model. \emph{African journal of business management} 4(19):4199.

\bibitem[{Yadlowsky et~al.(2024)Yadlowsky, Fleming, Shah, Brunskill, \protect\BIBand{} Wager}]{yadlowsky2024evaluating}
Yadlowsky S, Fleming S, Shah N, Brunskill E, Wager S (2024) Evaluating treatment prioritization rules via rank-weighted average treatment effects. \emph{Journal of the American Statistical Association} 1--14.

\bibitem[{Zhang et~al.(2012)Zhang, Tsiatis, Davidian, Zhang, \protect\BIBand{} Laber}]{zhang2012estimating}
Zhang B, Tsiatis AA, Davidian M, Zhang M, Laber E (2012) Estimating optimal treatment regimes from a classification perspective. \emph{Stat} 1(1):103--114.

\end{thebibliography}




\end{document}